\documentclass{article}
\usepackage{graphicx}

\usepackage[preprint]{neurips_2026}

\usepackage[utf8]{inputenc} 
\usepackage[T1]{fontenc}    
\usepackage{hyperref}       
\usepackage{url}            
\usepackage{booktabs}       
\usepackage{amsfonts}       
\usepackage{nicefrac}       
\usepackage{microtype}      
\usepackage{xcolor}         
\usepackage{algorithm}
\usepackage{algorithmic}
\usepackage{amsmath}
\usepackage{amssymb}
\usepackage{multirow}
\usepackage{caption}
\usepackage{wrapfig}
\usepackage{subcaption} 
\usepackage[table]{xcolor}

\title{Model Distribution-Aware Multimodal Dataset Distillation}

\author{%
  Jiajun Shen\thanks{Equal contribution.}
  \quad
  Yonglin Wu\footnotemark[1]
  \quad
  Ruonan Yu
  \quad
  Xinchao Wang\thanks{Corresponding author.} \\
  National University of Singapore \\
  \texttt{\{jiajun.shen,yonglin.wu,ruonan\}@u.nus.edu} \\
  \texttt{xinchao@nus.edu.sg}
}

\begin{document}

\maketitle

\begin{abstract}
Multimodal dataset distillation (MDD) aims to reduce the cost of vision-language pre-training by synthesizing compact image-text datasets. Although recent feature-matching methods have improved distillation efficiency, synthetic data may become specialized to the model representations encountered during optimization, limiting generalization to unseen model states. We propose \textbf{Model Distribution-Aware Multimodal Dataset Distillation (MDA-DD)}, which leverages supervision from a dynamic staggered queue of models at different training stages. We further introduce asymmetric dual projection for complementary alignment in visual and linguistic spaces and an uncentered second-order moment objective that retains mean-related cross-modal information. Experiments on Flickr30K and MS-COCO demonstrate consistent retrieval improvements over existing methods across synthetic-data budgets. With only 100 synthetic pairs, MDA-DD improves IR@1 and TR@1 on Flickr30K by 12.9 and 18.2 percentage points, respectively, over the strongest baseline at the same budget, outperforming all compared methods using 500 pairs on both datasets across all six retrieval metrics. Our code is available at \url{https://github.com/Jun-Shen04-Jia/MDA-DD}.
\end{abstract}

\section{Introduction}

\begin{wrapfigure}{r}{0.38\textwidth}
  \vspace{-15pt}
  \centering
  \includegraphics[trim=180 20 180 10, clip, width=\linewidth]{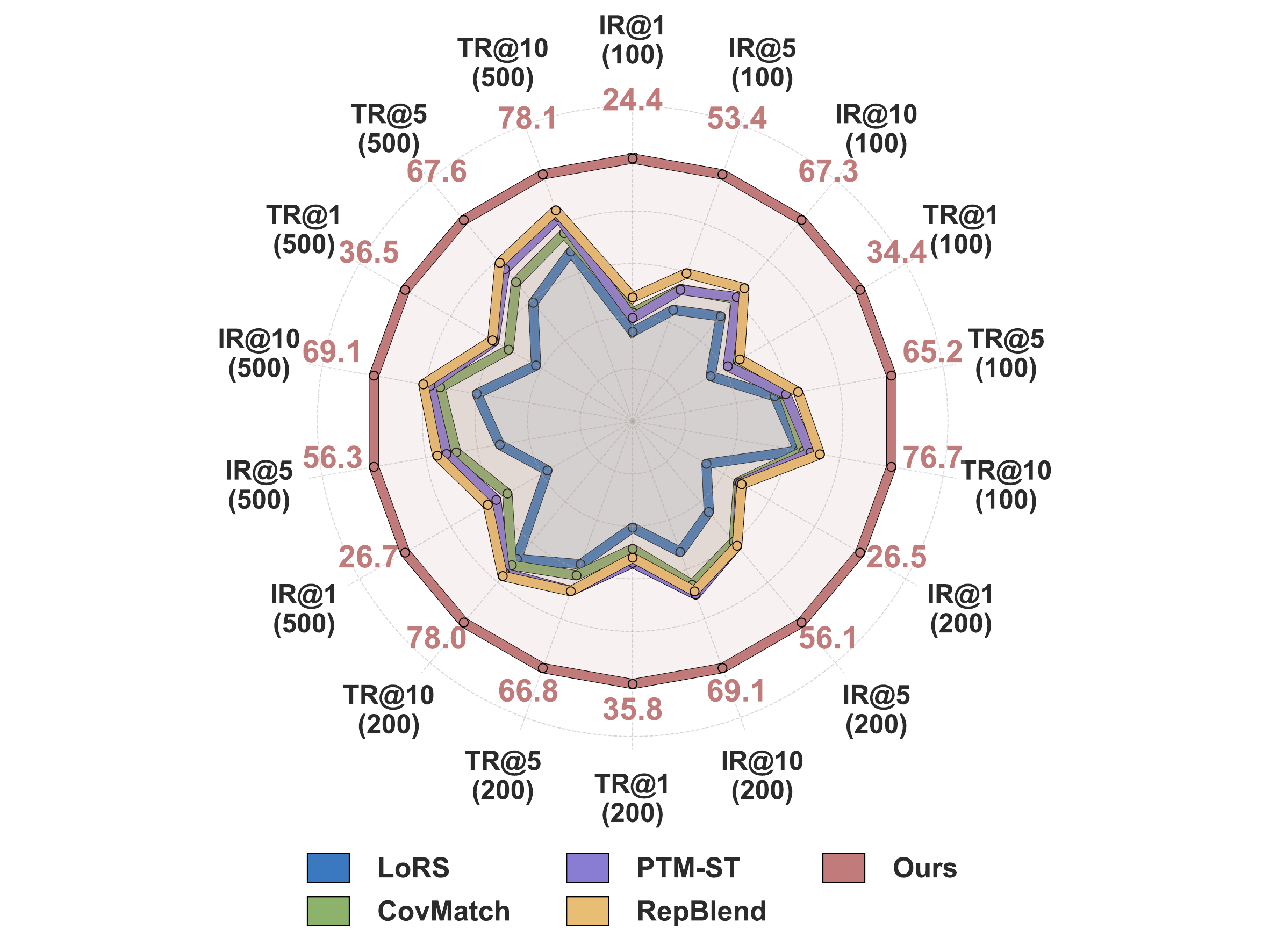}
  \caption{Radar charts comparing \textsc{MDA-DD} against all
  baselines on Flickr30K across retrieval metrics.}
  \label{fig:radar}
  \vspace{-25pt}
\end{wrapfigure}

Dataset distillation (DD) \cite{wang2018dataset,2026DeyuCVPR,2026LingaoDD,2026ODeyuICLR,2026RuonanFlashDD,2026OLetianICLR} aims to compress large datasets into compact synthetic training sets. Contrastive vision-language models such as CLIP \cite{radford2021learning} and SigLIP \cite{zhai2023sigmoid,tschannen2025siglip} rely heavily on web-scale image-text corpora, incurring substantial storage and computational costs. Although DD has proven effective across unimodal domains, extending it to image-text data requires preserving fine-grained semantic alignment between modalities.

Trajectory-based methods such as MTT-VL \cite{wuvision} and LoRS \cite{xu2024low} address this issue by matching expert training trajectories to guide cross-modal alignment. RepBlend \cite{zhangbeyond} further improves efficiency by freezing both encoders and matching projection-layer trajectories. However, generating expert trajectories still requires training on the full dataset and storing intermediate checkpoints, incurring substantial computational and storage costs. CovMatch \cite{leecovmatch} avoids expert trajectory generation by updating both encoders online and matching feature distributions between real and synthetic data, substantially reducing the cost of distillation.

\begin{figure}[t]
  \centering
\includegraphics[width=0.9\textwidth, trim=30 200 30 30 clip]{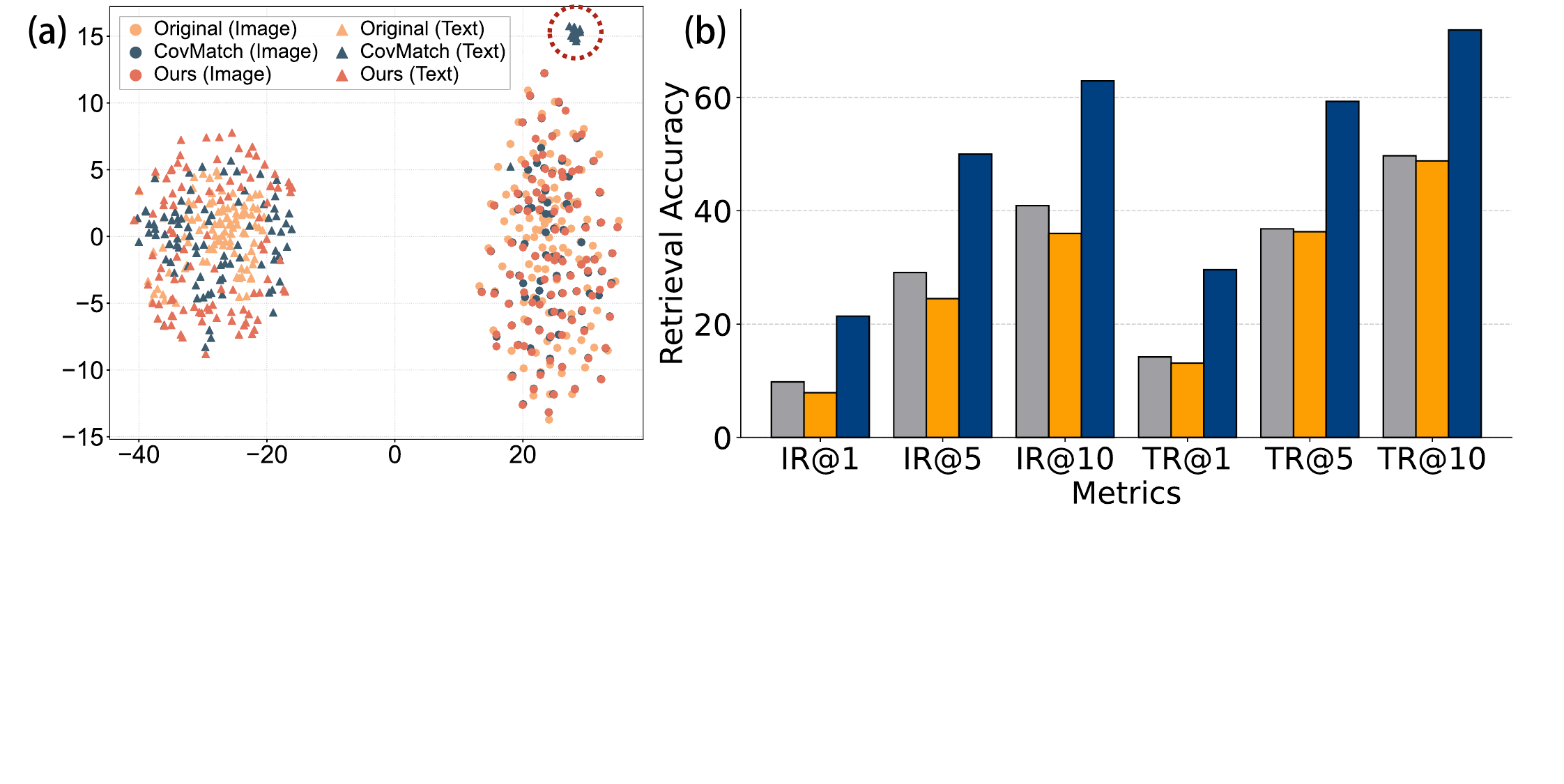}
  \caption{(a) Joint t-SNE visualization of multimodal embeddings under a different model initialization. Circles denote image embeddings and triangles denote text embeddings; colors indicate the source (\textcolor[HTML]{F9AE78}{\textbf{yellow:real data,}}  
 \textcolor[HTML]{3D5C6F}{\textbf{cyan:CovMatch,}} 
  \textcolor[HTML]{E47159}{\textbf{orange:MDA-DD}}). The \textcolor[HTML]{AF2518}{\textbf{red dashed circle}}  highlights a subset of CovMatch text embeddings that collapses to an isolated region. (b) Retrieval performance of CovMatch variants on Flickr30K with 100 pairs. Removing the image projection \textcolor[HTML]{FD9F02}{\textbf{(yellow)}} has minimal impact relative to the baseline \textcolor[HTML]{A0A0A4}{\textbf{(grey)}}, while random model initialization \textcolor[HTML]{004080}{\textbf{(blue)}} yields consistent gains across all metrics.}
\label{fig:TSNE}
\end{figure}

Despite this progress, feature matching remains dependent on the model representations used during distillation. CovMatch periodically resets its encoders to the same pretrained weights before updating them online, which may cause synthetic data to specialize to the encountered model states and degrade downstream retrieval performance under different initializations.  Figure~\ref{fig:TSNE}(a) shows a subset of its text embeddings separated from the main clusters under a different model initialization, while Figure~\ref{fig:TSNE}(b) shows that changing the initialization strategy improves retrieval performance.

Beyond model-state dependence, aligning both modalities in the visual feature space leaves complementary alignment in the linguistic space unexplored, while centered cross-covariance removes mean-related cross-modal information from the matched second-order statistic, potentially overlooking useful global cross-modal information.

We propose \textbf{Model Distribution-Aware Multimodal Dataset Distillation (MDA-DD)}, which uses a dynamic staggered model queue to provide supervision from models at different training stages. We further introduce \textbf{asymmetric dual projection (ADP)} for complementary alignment in visual and linguistic spaces and a \textbf{Second-Order Cross-Modal Moment (SOCM)} objective that retains mean-related cross-modal information without explicit centering.

Experiments on Flickr30K~\cite{plummer2015flickr30k} and MS-COCO~\cite{lin2014microsoft} show consistent improvements across synthetic-data budgets. With only 100 pairs, MDA-DD achieves 24.4 IR@1 and 34.4 TR@1 on Flickr30K, and 7.3 IR@1 and 10.5 TR@1 on MS-COCO. On both datasets, this 100-pair configuration outperforms all compared methods using 500 pairs across all six retrieval metrics.

Our main contributions are summarized as follows:

\textbf{(1)} We introduce a model-distribution-aware distillation framework with a dynamic staggered model queue that provides supervision from models at different training stages.

\textbf{(2)} We develop complementary cross-modal matching through asymmetric dual projection and an uncentered second-order moment objective that retains mean-related cross-modal information.

\textbf{(3)} We demonstrate substantial improvements in retrieval performance and data efficiency, as well as generalization to unseen visual encoder architectures.

\section{Preliminaries and Related Work}
\label{gen_inst}
We formalize multimodal dataset distillation for image-text contrastive learning and review existing distribution-based approaches (see Appendix \ref{app:trajectory} for a discussion on trajectory-based methods).

\paragraph{Multimodal Dataset Distillation.} Vision-language models embed image-text pairs into a joint representation space. Given a dataset $\mathcal{T}=\{(\mathbf{x}^{i}_{v},\,\mathbf{x}^{i}_{l})\}_{i=1}^{M}$, an image encoder $f_{v}$ and a text encoder $f_{l}$, each followed by a linear projection head, produce multimodal embeddings $\mathbf{z}_{v},\mathbf{z}_{l}\!\in\!\mathbb{R}^{z}$. The model is trained with the symmetric InfoNCE \cite{oord2018representation} loss: \begin{equation}\label{eq:infonce}
\mathcal{L}_{\mathrm{NCE}}(S) = -\frac{1}{2B}\sum_{i=1}^{B}\left[\log\frac{\exp(S_{ii}/\tau)}{\sum_{j=1}^{B}\exp(S_{ij}/\tau)}+\log\frac{\exp(S_{ii}/\tau)}{\sum_{j=1}^{B}\exp(S_{ji}/\tau)}\right].
\end{equation}
where $s_{ij}=\cos(\mathbf{z}^{i}_{v},\mathbf{z}^{j}_{l})$ and $\tau>0$ is a learnable temperature.

The goal of cross-modal dataset distillation is to compress
$\mathcal{T}$ into a compact synthetic set
$\mathcal{S}=\{(\tilde{\mathbf{x}}^{i}_{v},\,
\tilde{\mathbf{x}}^{i}_{l})\}_{i=1}^{N}$ with $N\!\ll\!M$.
Let $P_{\mathcal{D}}$ denote the distribution of the real data.
We seek the synthetic set such that a model trained on $\mathcal{S}$
attains cross-modal alignment comparable to training on the full
corpus $\mathcal{T}$, by minimizing the expected alignment gap
under $P_{\mathcal{D}}$:
\begin{equation}\label{eq:goal}
  \mathcal{S}^{*}
  \;=\;
  \mathop{\arg\min}\limits_{\mathcal{S}}\;
  \mathbb{E}_{(\mathbf{x}_{v},\mathbf{x}_{l})\sim P_{\mathcal{D}}}
  \bigl\|\,
    \mathcal{L}(\theta_{\mathcal{T}})
    \;-\;
    \mathcal{L}(\theta_{\mathcal{S}})
  \,\bigr\|,
\end{equation}
where $\theta_{\mathcal{T}}$ and $\theta_{\mathcal{S}}$ denote
model parameters obtained by training on $\mathcal{T}$ and
$\mathcal{S}$ respectively, and $\mathcal{L}$ is the cross-modal
retrieval loss.

\paragraph{Distribution-Based Paradigms.}
CovMatch replaces costly trajectory unrolling with single-step distribution matching. By interleaving model training with data distillation, CovMatch uses the updated model directly as a feature extractor for both real and synthetic sets. Its distillation objective combines a shared projection design with cross-covariance matching.

\smallskip\noindent \textbf{Projection Design.}
The image encoder extracts embeddings $\mathbf{e}_{v}\!\in\!\mathbb{R}^{d_{v}}$, which are mapped to $\mathbf{z}_{v}=\mathbf{W}_{v}^{\updownarrow}\mathbf{e}_{v}\!\in\!\mathbb{R}^{d_{v}}$  through a projection matrix $\mathbf{W}_{v}^{\updownarrow}\!\in\!\mathbb{R}^{d_{v}\times d_{v}}$. Similarly, the text encoder produces embeddings $\mathbf{e}_{l}\!\in\!\mathbb{R}^{d_{l}}$  and maps them to $\mathbf{z}_{l}=\mathbf{W}_{l}^{\uparrow}\mathbf{e}_{l}\!\in\!\mathbb{R}^{d_{v}}$ through a projection matrix $\mathbf{W}_{l}^{\uparrow}\!\in\!\mathbb{R}^{d_{v}\times d_{l}}$. This projection scheme aligns text representations with the visual feature space.

\smallskip\noindent \textbf{Cross-Covariance Matching.} 
Given a batch of $B$ paired embeddings $\{\mathbf{e}^{i}_{v},\mathbf{e}^{i}_{l}\}_{i=1}^{B}$, the batch means $\boldsymbol{\mu}_{v}$ and $\boldsymbol{\mu}_{l}$ are first computed to obtain centered embeddings $\bar{\mathbf{e}}^{i}_{v}=\mathbf{e}^{i}_{v}-\boldsymbol{\mu}_{v}$ and $\bar{\mathbf{e}}^{i}_{l}=\mathbf{e}^{i}_{l}-\boldsymbol{\mu}_{l}$. The cross-modal covariance matrix is then defined as:
\begin{equation}\label{eq:cov}
 \mathbf{C}
 \;=\;
 \frac{1}{B}\sum_{i=1}^{B}
 \bar{\mathbf{e}}^{i}_{v}\,
 (\bar{\mathbf{e}}^{i}_{l})^{\top}
 \;=\;
 \frac{1}{B}\,
 \bar{\mathbf{E}}_{v}^{\top}\,\bar{\mathbf{E}}_{l}
 \;\in\;\mathbb{R}^{d_{v}\times d_{l}},
\end{equation}
where $\bar{\mathbf{E}}_{v}\!\in\!\mathbb{R}^{B\times d_{v}}$ and $\bar{\mathbf{E}}_{l}\!\in\!\mathbb{R}^{B\times d_{l}}$ stack the centered embeddings row-wise.

The distillation loss aligns the cross-covariance matrices and the projected means of each modality: \begin{equation}\label{eq:covmatch}
 \mathcal{L}_{\mathrm{total}}
 \;=\;
 \bigl\|\mathbf{C}_{\mathcal{S}}
        -\rho\,\mathbf{C}_{\mathcal{T}}\bigr\|_{F}^{2}
 \;+\;
 \lambda\bigl(
   \|\bar{\mathbf{z}}_{v}^{\mathcal{S}}
    -\bar{\mathbf{z}}_{v}^{\mathcal{T}}\|_{2}^{2}
    +
   \|\bar{\mathbf{z}}_{l}^{\mathcal{S}}
    -\bar{\mathbf{z}}_{l}^{\mathcal{T}}\|_{2}^{2}
 \bigr),
\end{equation} where $\rho$ is a scaling factor, $\bar{\mathbf{z}}$ denotes the projected batch mean, and $\lambda$ is a balancing weight. This scheme avoids generating and storing expert trajectories.
\section{Method}
\label{sec:method}

\subsection{Model-Distribution-Aware Distillation}
\label{sec:mda}

\textbf{Motivation.} Feature-distribution matching depends on model representations that evolve during training, so synthetic data optimized along one trajectory may match real-data statistics less well under other model states. CovMatch~\cite{leecovmatch} introduces model-state variation through online training and periodic reinitialization, but its encoders repeatedly return to the same pretrained weights. Synthetic-data updates are also guided by one active model at a time, rather than by models at different training stages. This may cause synthetic data to specialize to the states encountered during distillation and generalize poorly to others.

To characterize this limitation, let $\mathcal{T}$ and $\mathcal{S}$ denote the real and synthetic datasets, respectively, and let $\mathcal{L}_{\mathrm{match}}(\theta;\mathcal{S},\mathcal{T})$ denote their feature-matching discrepancy under model state $\theta$, as defined in Eq.~\eqref{eq:total_loss}. Let $P_{\mathrm{seq}}(\Theta\mid\phi_0)$ represent the model-state distribution covered by sequential online training cycles starting from a common pretrained encoder initialization $\phi_0$, including their subsequent parameter updates. Let $P(\Theta)$ denote a broader target distribution of model states. We characterize potential model-state overfitting through the generalization gap
\begin{equation}
\label{eq:state_gap}
\begin{aligned}
\Delta_{\mathrm{state}}(\mathcal{S})
={}&
\mathbb{E}_{\theta\sim P(\Theta)}
\left[
\mathcal{L}_{\mathrm{match}}
(\theta;\mathcal{S},\mathcal{T})
\right]-
\mathbb{E}_{\theta\sim P_{\mathrm{seq}}(\Theta\mid\phi_0)}
\left[
\mathcal{L}_{\mathrm{match}}
(\theta;\mathcal{S},\mathcal{T})
\right].
\end{aligned}
\end{equation}
A positive \(\Delta_{\mathrm{state}}\) indicates that synthetic data matches real-data statistics more closely on the states covered during distillation than on the broader target distribution. This motivates expanding model-state coverage beyond the repeatedly encountered training trajectories.

\textbf{Distributional Objective.} To reduce this sensitivity, we formulate multimodal dataset distillation as feature matching over a distribution of model states:
\begin{equation}
\label{eq:dist_obj}
\begin{aligned}
\mathcal{S}^{*}
=\arg\min_{\mathcal{S}}\;
\mathbb{E}_{\theta\sim P(\Theta)}
\Big[
&\big\|
M_{\theta}(\mathcal{S})
-\rho M_{\theta}(\mathcal{T})
\big\|_F^2
+\lambda\big(
\|\bar z_{v,\theta}^{\mathcal{S}}
-\bar z_{v,\theta}^{\mathcal{T}}\|_2^2
+
\|\bar z_{l,\theta}^{\mathcal{S}}
-\bar z_{l,\theta}^{\mathcal{T}}\|_2^2
\big)
\Big].
\end{aligned}
\end{equation}
Here, $M_{\theta}(\cdot)$ denotes the second-order cross-modal moment defined in Eq.~\eqref{eq:socm_def}, while $\bar z_{v,\theta}^{\mathcal{D}}$ and $\bar z_{l,\theta}^{\mathcal{D}}$ are the mean projected features of the visual and linguistic modalities for dataset $\mathcal{D}$. The coefficient $\rho$ scales the real-data moment, and $\lambda$ balances moment matching and feature-mean regularization.

Since the target model-state distribution is generally inaccessible, we approximate its supervision using $K$ concurrently maintained online models at different training stages. At iteration $t$, the active queue $\mathcal{Q}_t=\{\theta_k\}_{k=1}^{K_t}$ induces the empirical matching objective
\begin{equation}
\label{eq:particle_obj}
\begin{aligned}
\widehat{\mathcal{L}}_{\mathrm{distill}}
(\mathcal{S};\mathcal{Q}_t)
=
\frac{1}{K_t}
\sum_{k=1}^{K_t}
\Big[
&\big\|
M_{\theta_k}(\mathcal{S})
-\rho M_{\theta_k}(\mathcal{T})
\big\|_F^2
+\lambda\big(
\|\bar z_{v,k}^{\mathcal{S}}
-\bar z_{v,k}^{\mathcal{T}}\|_2^2
+
\|\bar z_{l,k}^{\mathcal{S}}
-\bar z_{l,k}^{\mathcal{T}}\|_2^2
\big)
\Big],
\end{aligned}
\end{equation}
where \(K_t=|Q_t|\le K\) denotes the number of active models. The queue provides a stage-stratified approximation that covers different training states without treating its models as independent samples from the target distribution.

\begin{figure}[t]
  \centering
  \includegraphics[width=0.98\textwidth, trim=20 20 20 20, clip]{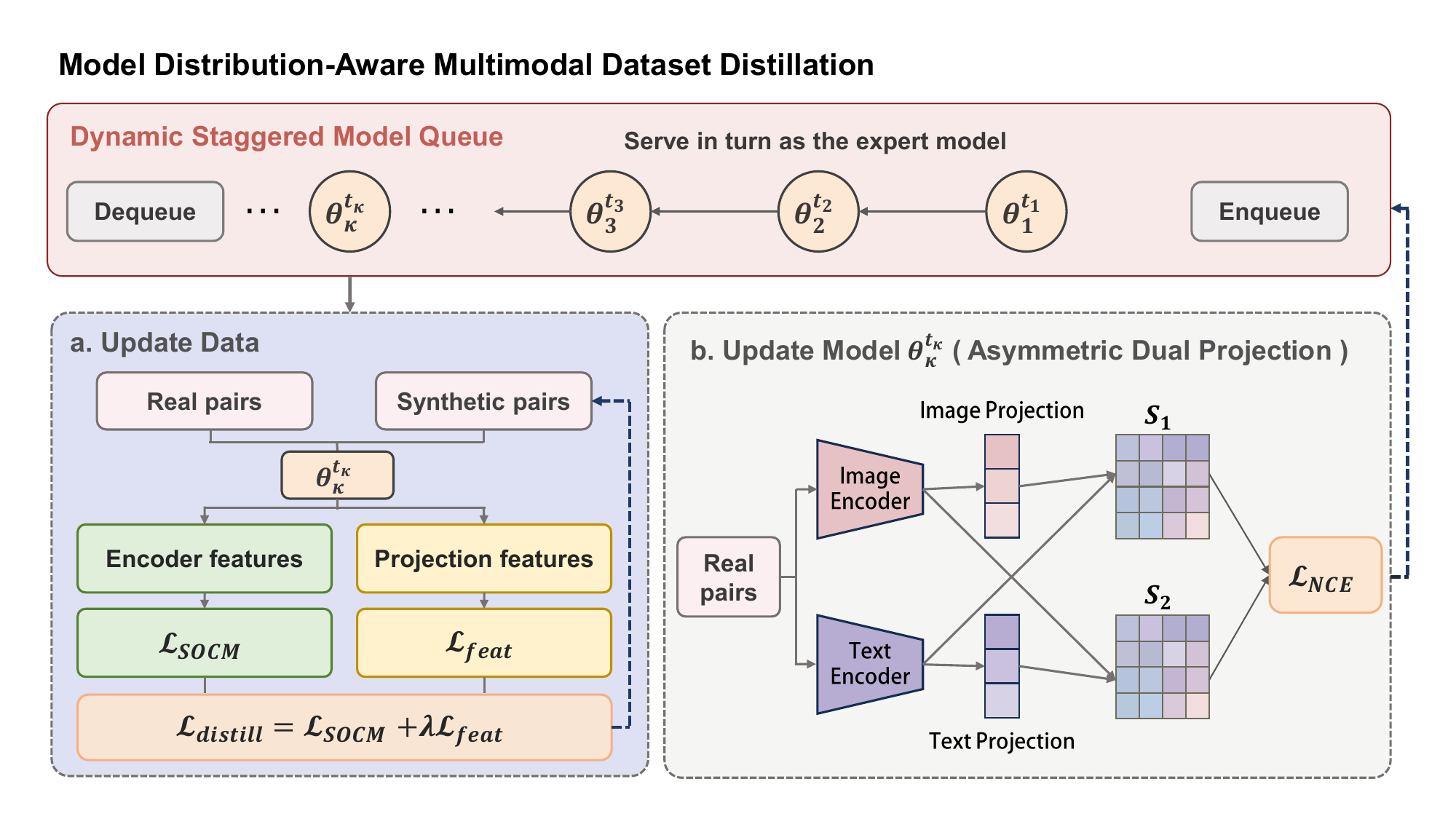}
 \caption{\textbf{Overview of MDA-DD.} A dynamic staggered model queue maintains expert models at different training stages. (a) \textbf{Update Data:} Synthetic pairs are optimized by matching encoder-level second-order cross-modal moments (SOCM) and projection-level feature means with real pairs. (b) \textbf{Update Model:} Each expert is trained on real pairs using asymmetric dual projections for bidirectional contrastive learning.}
  \label{fig:queue}
  \vspace{-5pt}
\end{figure}

\textbf{Dynamic Staggered Model Queue.} As illustrated in Figure~\ref{fig:queue}, we maintain a queue of at most $K$ online models, each with a maximum lifetime of $T_{\max}$ training steps on real data. New models share the pretrained encoder initialization but use independently initialized projection heads and separate training randomness. Each model evolves through online contrastive training, with its age defined by the number of real-data updates since initialization.

To maintain coverage across training stages, we introduce new models at intervals of $\Delta t=T_{\max}/K$ and retire the oldest model when it reaches its maximum lifetime. During the initial warm-up period, the queue is gradually populated. Once the queue reaches its steady state, the active models span different training ages. Let $a_{(k)}(t)$ denote the age of the $k$-th model ordered from youngest to oldest. The steady-state ages satisfy
\begin{equation}
\label{eq:queue_age}
\begin{aligned}
\mathcal{Q}_t
&=\{\theta_k^{(t)}\}_{k=1}^{K},
\qquad
a_{(k)}(t)
=(k-1)\Delta t+t',
\qquad
t'=t\bmod\Delta t,
\end{aligned}
\end{equation}
where the iteration index is aligned with the model replacement cycle. Consecutive models are therefore separated by $\Delta t$ training steps, providing simultaneous supervision from early, intermediate, and later stages of model optimization.

At each distillation iteration, the active models take turns updating the synthetic dataset. For each model $\theta_k$, we first extract encoder-level cross-modal moments and projection-level feature means from real and synthetic image-text batches. Keeping $\theta_k$ fixed, we treat the real-data statistics as targets and backpropagate the matching loss in Eq.~\eqref{eq:total_loss} only through the synthetic branch to update $\mathcal{S}$. The model is then advanced by one contrastive-learning step on real data using the asymmetric dual projection objective (\S\ref{sec:projection}). After all active models have been processed, the queue is updated by retiring expired models and introducing new ones according to the staggered schedule.

Following Algorithm~\ref{alg:mda-dd}, synthetic-data updates are performed sequentially across the active models, providing an incremental optimization of the multi-state objective in Eq.~\eqref{eq:particle_obj}. By exposing synthetic pairs to multiple evolving feature geometries within each distillation iteration, the dynamic queue broadens model-state supervision without requiring stored expert trajectories.

\subsection{Asymmetric Dual Projection}
\label{sec:projection}

\paragraph{Limitations of Existing Projection Design.}
CovMatch aligns both modalities in the visual feature space $\mathbb{R}^{d_v}$ by projecting image embeddings with $W_v \in \mathbb{R}^{d_v \times d_v}$ and text embeddings with $W_l \in \mathbb{R}^{d_v \times d_l}$. These projections produce a single similarity matrix $S$ for contrastive learning. This design has two limitations: the same-dimensional image projection $W_v$ may be absorbed into the trainable image encoder's final layer, while text embeddings are projected into the visual space without a corresponding projection in the reverse direction. Consequently, cross-modal alignment is restricted to the visual space, leaving complementary alignment in the linguistic space unexplored.

\paragraph{Dual Projection.}
 We replace the redundant same-dimension image projection with two asymmetric cross-dimensional matrices, $\mathbf{W}{v}^{\downarrow}\!\in\!\mathbb{R}^{d_l\times d_v}$ and $\mathbf{W}{l}^{\uparrow}\!\in\!\mathbb{R}^{d_v\times d_l}$, which project image and text embeddings into the feature space of the other modality:
\begin{equation}\label{eq:dual_proj}
  \mathbf{z}_{v}
  = \mathbf{W}_{v}^{\downarrow}\,\mathbf{e}_{v}
  \in\mathbb{R}^{d_l},
  \qquad
  \mathbf{z}_{l}
  = \mathbf{W}_{l}^{\uparrow}\,\mathbf{e}_{l}
  \in\mathbb{R}^{d_v}.
\end{equation}
Neither projection is redundant: $\mathbf{W}_{v}^{\downarrow}$
genuinely reduces dimensionality and cannot be absorbed by the image
encoder, while $\mathbf{W}_{l}^{\uparrow}$ performs the complementary
expansion.

\paragraph{Dual-Space Similarity.}
The two projected representations produce two cross-modal similarity matrices in different feature spaces. Within the text space $\mathbb{R}^{d_l}$, projected image embeddings are compared with original text embeddings, whereas within the visual space $\mathbb{R}^{d_v}$, original image embeddings are compared with projected text embeddings:
\begin{equation}\label{eq:dual_sim}
  \mathbf{S}_{1}^{(ij)}
  = \cos\!\bigl (\mathbf{z}_{v}^{i},\;\mathbf{e}_{l}^{j}\bigr),
  \qquad
  \mathbf{S}_{2}^{(ij)}
  = \cos\!\bigl (\mathbf{e}_{v}^{i},\;\mathbf{z}_{l}^{j}\bigr).
\end{equation}
$\mathbf{S}_{1}$ quantifies how effectively visual content is represented in the linguistic space, whereas $\mathbf{S}_{2}$ quantifies how reliably textual content is grounded in the visual space. Together, they provide complementary alignment signals through the two projection directions, rather than relying on a single feature space. We then apply the symmetric InfoNCE loss in Eq.~\ref{eq:infonce} to each similarity matrix and average the resulting losses.

\begin{algorithm}[t]
\caption{\textsc{MDA-DD}: Model-Distribution-Aware Multimodal Dataset Distillation}
\label{alg:mda-dd}
\begin{algorithmic}[1]
\REQUIRE Real dataset $\mathcal{T}$, synthetic size $N$, iterations $T_{\mathrm{iter}}$, queue size $K$, max lifetime $T_{\max}$, learning rates $\eta_{\mathcal{S}},\eta_{\theta}$
\ENSURE Synthetic dataset $\mathcal{S}=\{(\tilde{\mathbf{x}}_{v}^{i},\,\tilde{\mathbf{x}}_{l}^{i})\}_{i=1}^{N}$
\STATE Initialize $\mathcal{S}$ by sampling $N$ pairs from $\mathcal{T}$
\STATE Initialize queue $\mathcal{Q}=\{\theta_1\}$ with $t_1=0$ \COMMENT{$t_k$: training steps of $\theta_k$}
\FOR{$t = 1$ \TO $T_{\mathrm{iter}}$}
    \FOR{each $\theta_k \in \mathcal{Q}$}
        \STATE \textit{// Compute real-data statistics (frozen)}
        \STATE Sample $\mathcal{B}\!\sim\!\mathcal{T}$; compute $\mathbf{M}_{\theta_k}(\mathcal{B})$ via Eq.~\ref{eq:socm_def} and $\bar{\mathbf{z}}_{v}^{\,\mathcal{T}},\bar{\mathbf{z}}_{l}^{\,\mathcal{T}}$
        \STATE \textit{// Compute synthetic-data statistics (with gradients)}
        \STATE Sample $\tilde{\mathcal{B}}\!\sim\!\mathcal{S}$; compute $\mathbf{M}_{\theta_k}(\tilde{\mathcal{B}})$ and $\bar{\mathbf{z}}_{v}^{\,\mathcal{S}},\bar{\mathbf{z}}_{l}^{\,\mathcal{S}}$
        \STATE \textit{// Update synthetic data}
        \STATE Compute $\mathcal{L}_{\mathrm{total}}$ via Eq.~\ref{eq:total_loss}; \; $\mathcal{S}\leftarrow\mathcal{S}-\eta_{\mathcal{S}}\nabla_{\mathcal{S}}\mathcal{L}_{\mathrm{total}}$
        \STATE \textit{// Advance model on real data}
        \STATE Sample $\mathcal{B}'\!\sim\!\mathcal{T}$; \; $\theta_k\leftarrow\theta_k-\eta_{\theta}\nabla_{\theta_k}\mathcal{L}_{\mathrm{NCE}}(\theta_k,\mathcal{B}')$; \; $t_k\leftarrow t_k+1$
    \ENDFOR
    \STATE \textit{// Queue maintenance}
    \IF{$\max_k t_k \geq T_{\max}$}
        \STATE Remove oldest $\theta_k$ from $\mathcal{Q}$
    \ENDIF
    \IF{$|\mathcal{Q}|<K$ \AND $t \bmod \lfloor T_{\max}/K\rfloor = 0$}
        \STATE Enqueue freshly initialized $\theta_{\mathrm{new}}$ with $t_{\mathrm{new}}=0$
    \ENDIF
\ENDFOR
\end{algorithmic}
\end{algorithm}

\subsection{Second-Order Cross-Modal Moment}
\label{sec:socm}

\paragraph{From Covariance to Raw Moment.}
The cross-modal covariance in CovMatch (Eq.~\ref{eq:cov}) requires explicit mean centering before computing the outer product. We instead match the raw second-order cross-modal moment, avoiding explicit centering.

Given a dataset
$\mathcal{D}=\{(\mathbf{x}_v^i,\,\mathbf{x}_l^i)\}_{i=1}^{n}$ and
model parameters $\theta$, let
$\mathbf{e}_v^i=f_v(\mathbf{x}_v^i;\theta_v)\!\in\!\mathbb{R}^{d_v}$
and
$\mathbf{e}_l^i=f_l(\mathbf{x}_l^i;\theta_l)\!\in\!\mathbb{R}^{d_l}$
denote the encoder outputs. The Second-Order Cross-Modal
Moment (SOCM) is defined as:
\begin{equation}\label{eq:socm_def}
  \mathbf{M}_\theta(\mathcal{D})
  \;\triangleq\;
  \frac{1}{n}\sum_{i=1}^{n}
  \mathbf{e}_v^i\,{\mathbf{e}_l^i}^{\!\top}
  \;=\;
  \frac{1}{n}\,\mathbf{E}_v^{\top}\,\mathbf{E}_l
  \;\in\;\mathbb{R}^{d_v\times d_l},
\end{equation}
where
$\mathbf{E}_v\!\in\!\mathbb{R}^{n\times d_v}$ and
$\mathbf{E}_l\!\in\!\mathbb{R}^{n\times d_l}$ stack the
encoder outputs row-wise. Its relationship to the cross-modal
covariance $\mathbf{C}$ (Eq.~\ref{eq:cov}) is:
\begin{equation}\label{eq:socm_decomp}
\mathbf{M}_\theta(\mathcal{D})\;=\;\underbrace{\mathbf{C}}_{\text{covariance}}\;+\;\underbrace{\boldsymbol{\mu}_v\,\boldsymbol{\mu}_l^{\!\top}
  }_{\text{mean outer product}},
\end{equation}

where $\boldsymbol{\mu}_v$ and $\boldsymbol{\mu}_l$ denote the batch means. SOCM therefore captures both the centered cross-modal covariance and the outer product of the modality means within a single matrix. Computing the raw moment directly avoids the mean-centering and element-wise subtraction required by covariance matching, reducing the computational cost of the second-order matching term.

\begin{table}[t]
  \definecolor{ourshl}{RGB}{235,225,250}
  \centering
  \caption{
    Cross-modal retrieval on \textbf{Flickr30K} (NFNet+BERT).
    The model trained on full dataset achieves:
    IR@1 = 47.8, IR@5 = 79.0, IR@10 = 87.4,
    TR@1 = 61.5, TR@5 = 85.1, TR@10 = 92.1.
  }
  \label{tab:flickr}
  \small
  \setlength{\tabcolsep}{3.5pt}
  \begin{tabular}{lll cccc ccccc}
    \toprule
    \multirow{2}{*}{Pairs} & \multirow{2}{*}{Ratio} & \multirow{2}{*}{Metric} 
      & \multicolumn{4}{c}{Coreset selection} 
      & \multicolumn{5}{c}{Dataset distillation} \\
    \cmidrule(r){4-7} \cmidrule(l){8-12}
     &  &  & Rand & Herd & K-Cent & Forget & LoRS & CovMatch & RepBlend & PTM-ST & \cellcolor{ourshl}\textbf{Ours} \\
    \midrule
    \multirow{6}{*}{100} & \multirow{6}{*}{0.3\%} 
      & IR@1  & 1.0 & 0.7 & 0.7 & 0.7 & 8.3{\tiny$\pm$0.2}  & 10.1{\tiny$\pm$0.2} & 11.5{\tiny$\pm$0.4} & 9.6{\tiny$\pm$0.3}  & \cellcolor{ourshl}\textbf{24.4{\tiny$\pm$0.6}} \\
     &  & IR@5  & 4.0 & 2.8 & 3.1 & 2.4 & 24.1{\tiny$\pm$0.2} & 28.6{\tiny$\pm$0.4} & 32.0{\tiny$\pm$0.7} & 28.4{\tiny$\pm$0.5} & \cellcolor{ourshl}\textbf{53.4{\tiny$\pm$0.7}} \\
     &  & IR@10 & 6.5 & 5.3 & 6.1 & 5.6 & 35.1{\tiny$\pm$0.3} & 40.9{\tiny$\pm$0.6} & 44.5{\tiny$\pm$0.6} & 41.5{\tiny$\pm$0.5} & \cellcolor{ourshl}\textbf{67.3{\tiny$\pm$0.7}} \\
     &  & TR@1  & 1.3 & 1.1 & 0.6 & 1.2 & 11.8{\tiny$\pm$0.2} & 14.8{\tiny$\pm$0.9} & 16.2{\tiny$\pm$0.8} & 14.4{\tiny$\pm$0.4} & \cellcolor{ourshl}\textbf{34.4{\tiny$\pm$0.7}} \\
     &  & TR@5  & 5.9 & 4.7 & 5.0 & 4.2 & 35.8{\tiny$\pm$0.6} & 38.0{\tiny$\pm$0.4} & 41.7{\tiny$\pm$0.9} & 38.6{\tiny$\pm$0.8} & \cellcolor{ourshl}\textbf{65.2{\tiny$\pm$0.9}} \\
     &  & TR@10 & 10.1 & 7.9 & 7.6 & 9.7 & 49.2{\tiny$\pm$0.5} & 50.6{\tiny$\pm$0.6} & 55.5{\tiny$\pm$0.4} & 52.7{\tiny$\pm$0.7} & \cellcolor{ourshl}\textbf{76.7{\tiny$\pm$0.7}} \\
    \midrule
    \multirow{6}{*}{200} & \multirow{6}{*}{0.7\%} 
      & IR@1  & 1.1 & 1.5 & 1.5 & 1.2 & 8.6{\tiny$\pm$0.3}  & 12.3{\tiny$\pm$0.4} & 12.7{\tiny$\pm$0.8} & 12.5{\tiny$\pm$0.1} & \cellcolor{ourshl}\textbf{26.5{\tiny$\pm$0.5}} \\
     &  & IR@5  & 4.8 & 5.5 & 5.4 & 3.1 & 25.3{\tiny$\pm$0.2} & 33.6{\tiny$\pm$0.3} & 34.7{\tiny$\pm$0.6} & 34.7{\tiny$\pm$0.2} & \cellcolor{ourshl}\textbf{56.1{\tiny$\pm$0.6}} \\
     &  & IR@10 & 9.2 & 9.3 & 9.9 & 8.4 & 36.6{\tiny$\pm$0.3} & 45.8{\tiny$\pm$0.2} & 47.6{\tiny$\pm$0.5} & 48.5{\tiny$\pm$0.2} & \cellcolor{ourshl}\textbf{69.1{\tiny$\pm$0.7}} \\
     &  & TR@1  & 2.1 & 2.3 & 2.2 & 1.5 & 14.5{\tiny$\pm$0.5} & 17.4{\tiny$\pm$0.5} & 18.6{\tiny$\pm$0.7} & 19.3{\tiny$\pm$0.2} & \cellcolor{ourshl}\textbf{35.8{\tiny$\pm$0.8}} \\
     &  & TR@5  & 8.7 & 8.4 & 8.2 & 8.4 & 38.7{\tiny$\pm$0.5} & 41.7{\tiny$\pm$0.5} & 46.0{\tiny$\pm$0.8} & 45.9{\tiny$\pm$0.4} & \cellcolor{ourshl}\textbf{66.8{\tiny$\pm$0.6}} \\
     &  & TR@10 & 13.2 & 14.4 & 13.5 & 10.2 & 53.4{\tiny$\pm$0.5} & 55.8{\tiny$\pm$0.5} & 60.0{\tiny$\pm$0.6} & 58.9{\tiny$\pm$0.5} & \cellcolor{ourshl}\textbf{78.0{\tiny$\pm$0.8}} \\
    \midrule
    \multirow{6}{*}{500} & \multirow{6}{*}{1.7\%} 
      & IR@1  & 2.4  & 3.0  & 3.5  & 1.8  & 10.0{\tiny$\pm$0.2} & 14.7{\tiny$\pm$0.3} & 17.0{\tiny$\pm$0.6} & 16.0{\tiny$\pm$0.2} & \cellcolor{ourshl}\textbf{26.7{\tiny$\pm$1.2}} \\
     &  & IR@5  & 10.5 & 10.0 & 10.4 & 9.0  & 28.9{\tiny$\pm$0.7} & 38.4{\tiny$\pm$0.4} & 42.5{\tiny$\pm$0.5} & 40.5{\tiny$\pm$0.3} & \cellcolor{ourshl}\textbf{56.3{\tiny$\pm$1.5}} \\
     &  & IR@10 & 17.4 & 17.0 & 17.3 & 15.9 & 41.6{\tiny$\pm$0.6} & 51.4{\tiny$\pm$0.3} & 55.9{\tiny$\pm$0.6} & 54.0{\tiny$\pm$0.3} & \cellcolor{ourshl}\textbf{69.1{\tiny$\pm$1.2}} \\
     &  & TR@1  & 5.2  & 5.1  & 4.9  & 3.6  & 15.5{\tiny$\pm$0.7} & 19.9{\tiny$\pm$0.6} & 22.5{\tiny$\pm$0.4} & 22.2{\tiny$\pm$0.6} & \cellcolor{ourshl}\textbf{36.5{\tiny$\pm$1.1}} \\
     &  & TR@5  & 18.3 & 16.4 & 16.4 & 12.3 & 39.8{\tiny$\pm$0.4} & 46.7{\tiny$\pm$0.9} & 53.2{\tiny$\pm$0.3} & 51.1{\tiny$\pm$0.3} & \cellcolor{ourshl}\textbf{67.6{\tiny$\pm$1.3}} \\
     &  & TR@10 & 25.7 & 24.3 & 23.3 & 19.3 & 53.7{\tiny$\pm$0.3} & 59.5{\tiny$\pm$0.7} & 66.7{\tiny$\pm$0.3} & 64.6{\tiny$\pm$0.3} & \cellcolor{ourshl}\textbf{78.1{\tiny$\pm$1.0}} \\
    \bottomrule
  \end{tabular}
\end{table}
\paragraph{Complete Distillation Loss.}
We pair SOCM matching with a per-modality mean constraint at the
projection level. The full objective is:
\begin{equation}\label{eq:total_loss}
\begin{aligned}
  \mathcal{L}_{\mathrm{total}}
  \;=\;&
  \underbrace{
    \bigl\|\mathbf{M}_\theta(\mathcal{S})
           -\rho\,\mathbf{M}_\theta(\mathcal{T})\bigr\|_F^2
  }_{\mathcal{L}_{\mathrm{SOCM}}}
  + \lambda\underbrace{
    \Bigl(
      \bigl\|\bar{\mathbf{z}}_{v}^{\,\mathcal{S}}
             -\bar{\mathbf{z}}_{v}^{\,\mathcal{T}}\bigr\|_2^2
      +
      \bigl\|\bar{\mathbf{z}}_{l}^{\,\mathcal{S}}
             -\bar{\mathbf{z}}_{l}^{\,\mathcal{T}}\bigr\|_2^2
    \Bigr)
  }_{\mathcal{L}_{\mathrm{feat}}},
\end{aligned}
\end{equation}

where $\bar{\mathbf{z}}_{v}^{\,\mathcal{D}}$ and $\bar{\mathbf{z}}_{l}^{\,\mathcal{D}}$ are the batch-wise mean projected features, $\rho$ is a scaling coefficient, and $\lambda$ balances the two terms. $\mathcal{L}_{\mathrm{SOCM}}$ captures second-order cross-modal statistics at the encoder level, while $\mathcal{L}_{\mathrm{feat}}$ matches the mean features of each modality at the projection level.

\section{Experiments}
\label{sec:experiments}

\subsection{Experimental Setup}
\label{sec:setup}

We evaluate on two standard vision-language benchmarks: Flickr30K \cite{plummer2015flickr30k} (${\sim}$31K images) and MS-COCO \cite{lin2014microsoft} (${\sim}$123K images), each with five human-annotated captions per image. Using the Karpathy split \cite{karpathy2015deep}, Flickr30K follows a 29K/1K/1K train/val/test partition, while MS-COCO follows 113K/5K/5K, with results reported on the full 5K test set. Cross-modal candidates are ranked by cosine similarity  within the shared embedding space. We report Recall@$K$ ($K\!=\!1,5,10$) for both image-to-text retrieval (TR@$K$) and text-to-image retrieval (IR@$K$). Following prior work \cite{wuvision,xu2024low}, we use an ImageNet-pretrained NFNet~\cite{brock2021high} as the image encoder and a pretrained BERT-base \cite{devlin2019bert} as the text encoder. Each encoder is equipped with the asymmetric dual projection layer (\S\ref{sec:projection}) to map features into two complementary shared spaces for bidirectional alignment.
\paragraph{Baselines.}
We compare with two families of methods. (i) Coreset selection: Random Sampling, Herding \cite{welling2009herding}, K-Center \cite{farahani2009facility}, and Forgetting \cite{tonevaempirical}, all adapted to the vision-language setting following prior work. (ii) Dataset distillation: LoRS \cite{xu2024low}, RepBlend \cite{zhangbeyond}, CovMatch \cite{leecovmatch}, and PTM-ST \cite{guomultimodal}.

\paragraph{Implementation Details.}
We report the mean and standard deviation across 5 independent runs. All experiments are conducted on a single NVIDIA RTX 5090 (32~GB). Detailed hyperparameters and optimization settings are provided in the appendix.

\subsection{Main Results}
\label{sec:main_results}

Tables~\ref{tab:flickr} and~\ref{tab:coco} report cross-modal retrieval results on Flickr30K and MS-COCO under synthetic budgets of 100, 200, and 500 pairs. \textsc{MDA-DD} ranks first on every budget--metric combination on both benchmarks. At 100 pairs, \textsc{MDA-DD} reaches 24.4 IR@1 and 34.4 TR@1 on Flickr30K (0.3\% of training data) and 7.3 IR@1 and 10.5 TR@1 on MS-COCO (0.8‰), more than doubling the strongest baseline on Flickr30K and outperforming it by +3.2 IR@1 and +5.3 TR@1 on MS-COCO. On both datasets, this 100-pair configuration already surpasses all baselines operating at 500 pairs, yielding an effective 5$\times$ reduction in synthetic data requirement.

\begin{table}[t]
  \definecolor{ourshl}{RGB}{235,225,250}
  \centering
  \caption{
    Cross-modal retrieval on \textbf{MS-COCO} (NFNet+BERT).
    The model trained on full dataset achieves:
    IR@1 = 24.9, IR@5 = 53.7, IR@10 = 67.5,
    TR@1 = 32.5, TR@5 = 61.9, TR@10 = 74.0.
  }
  \label{tab:coco}
  \small
  \setlength{\tabcolsep}{3.5pt}
  \begin{tabular}{lll cccc ccccc}
    \toprule
    \multirow{2}{*}{Pairs} & \multirow{2}{*}{Ratio} & \multirow{2}{*}{Metric}
      & \multicolumn{4}{c}{Coreset selection}
      & \multicolumn{5}{c}{Dataset distillation} \\
    \cmidrule(r){4-7} \cmidrule(l){8-12}
     & & & Rand & Herd & K-Cent & Forget & LoRS & CovMatch & RepBlend & PTM-ST & \cellcolor{ourshl}\textbf{Ours} \\
    \midrule
    \multirow{6}{*}{100} & \multirow{6}{*}{0.8\textperthousand}
      & IR@1  & 0.3 & 0.5 & 0.4 & 0.3 & 1.8{\tiny$\pm$0.1} & 2.8{\tiny$\pm$0.1}  & 4.1{\tiny$\pm$0.3}  & 2.3{\tiny$\pm$0.1}  & \cellcolor{ourshl}\textbf{7.3{\tiny$\pm$0.2}}  \\
     & & IR@5  & 1.3 & 1.4 & 1.4 & 1.5 & 7.1{\tiny$\pm$0.2} & 10.5{\tiny$\pm$0.2} & 13.9{\tiny$\pm$0.8} & 9.0{\tiny$\pm$0.1}  & \cellcolor{ourshl}\textbf{22.6{\tiny$\pm$0.4}} \\
     & & IR@10 & 2.7 & 3.5 & 2.5 & 2.5 & 12.2{\tiny$\pm$0.2} & 17.7{\tiny$\pm$0.3} & 22.3{\tiny$\pm$0.5} & 15.6{\tiny$\pm$0.3} & \cellcolor{ourshl}\textbf{34.4{\tiny$\pm$0.6}} \\
     & & TR@1  & 0.8 & 0.8 & 1.4 & 0.7 & 3.3{\tiny$\pm$0.2} & 3.8{\tiny$\pm$0.1}  & 5.2{\tiny$\pm$0.5}  & 4.1{\tiny$\pm$0.2}  & \cellcolor{ourshl}\textbf{10.5{\tiny$\pm$0.2}} \\
     & & TR@5  & 3.0 & 2.1 & 3.7 & 2.6 & 12.2{\tiny$\pm$0.3} & 13.1{\tiny$\pm$0.3} & 17.9{\tiny$\pm$0.9} & 13.4{\tiny$\pm$0.2} & \cellcolor{ourshl}\textbf{28.6{\tiny$\pm$0.6}} \\
     & & TR@10 & 5.0 & 4.9 & 5.5 & 4.8 & 19.6{\tiny$\pm$0.3} & 21.1{\tiny$\pm$0.2} & 28.0{\tiny$\pm$0.3} & 22.0{\tiny$\pm$0.2} & \cellcolor{ourshl}\textbf{41.8{\tiny$\pm$0.7}} \\
    \midrule
    \multirow{6}{*}{200} & \multirow{6}{*}{1.7\textperthousand}
      & IR@1  & 0.6 & 0.9 & 0.7 & 0.6 & 2.4{\tiny$\pm$0.1} & 3.8{\tiny$\pm$0.1}  & 6.1{\tiny$\pm$0.8}  & 3.9{\tiny$\pm$0.1}  & \cellcolor{ourshl}\textbf{8.3{\tiny$\pm$0.2}}  \\
     & & IR@5  & 2.3 & 2.4 & 2.1 & 2.8 & 9.3{\tiny$\pm$0.2} & 13.4{\tiny$\pm$0.1} & 19.3{\tiny$\pm$0.7} & 13.7{\tiny$\pm$0.2} & \cellcolor{ourshl}\textbf{25.6{\tiny$\pm$0.4}} \\
     & & IR@10 & 4.4 & 4.1 & 5.8 & 4.9 & 15.5{\tiny$\pm$0.2} & 21.8{\tiny$\pm$0.2} & 29.8{\tiny$\pm$0.5} & 22.2{\tiny$\pm$0.2} & \cellcolor{ourshl}\textbf{38.4{\tiny$\pm$0.5}} \\
     & & TR@1  & 1.0 & 1.0 & 1.2 & 1.1 & 4.3{\tiny$\pm$0.1} & 5.3{\tiny$\pm$0.2}  & 6.9{\tiny$\pm$0.6}  & 5.7{\tiny$\pm$0.2}  & \cellcolor{ourshl}\textbf{12.1{\tiny$\pm$0.3}} \\
     & & TR@5  & 4.0 & 3.6 & 3.8 & 3.5 & 14.2{\tiny$\pm$0.3} & 17.3{\tiny$\pm$0.2} & 21.8{\tiny$\pm$0.9} & 18.2{\tiny$\pm$0.3} & \cellcolor{ourshl}\textbf{32.6{\tiny$\pm$0.5}} \\
     & & TR@10 & 7.2 & 7.7 & 7.5 & 7.0 & 22.6{\tiny$\pm$0.2} & 27.0{\tiny$\pm$0.2} & 32.3{\tiny$\pm$0.7} & 27.8{\tiny$\pm$0.3} & \cellcolor{ourshl}\textbf{45.6{\tiny$\pm$0.8}} \\
    \midrule
    \multirow{6}{*}{500} & \multirow{6}{*}{4.4\textperthousand}
      & IR@1  & 1.1  & 1.7  & 1.1  & 0.8  & 2.8{\tiny$\pm$0.2}  & 5.4{\tiny$\pm$0.1}  & 6.2{\tiny$\pm$0.1}  & 6.6{\tiny$\pm$0.1}  & \cellcolor{ourshl}\textbf{9.5{\tiny$\pm$0.4}}  \\
     & & IR@5  & 5.0  & 5.3  & 6.3  & 5.8  & 9.9{\tiny$\pm$0.5}  & 18.0{\tiny$\pm$0.1} & 19.9{\tiny$\pm$0.3} & 20.5{\tiny$\pm$0.2} & \cellcolor{ourshl}\textbf{27.5{\tiny$\pm$0.7}} \\
     & & IR@10 & 8.7  & 9.9  & 10.5 & 8.2  & 16.5{\tiny$\pm$0.7} & 28.2{\tiny$\pm$0.1} & 30.6{\tiny$\pm$0.1} & 30.7{\tiny$\pm$0.2} & \cellcolor{ourshl}\textbf{40.3{\tiny$\pm$0.8}} \\
     & & TR@1  & 1.9  & 1.9  & 2.5  & 2.1  & 5.3{\tiny$\pm$0.5}  & 8.1{\tiny$\pm$0.3}  & 7.0{\tiny$\pm$0.2}  & 6.9{\tiny$\pm$0.3}  & \cellcolor{ourshl}\textbf{13.0{\tiny$\pm$0.6}} \\
     & & TR@5  & 7.5  & 7.8  & 8.7  & 8.2  & 18.3{\tiny$\pm$1.5} & 23.5{\tiny$\pm$0.3} & 22.0{\tiny$\pm$0.3} & 20.1{\tiny$\pm$0.2} & \cellcolor{ourshl}\textbf{34.7{\tiny$\pm$1.1}} \\
     & & TR@10 & 12.5 & 13.7 & 14.3 & 13.0 & 27.9{\tiny$\pm$1.4} & 34.6{\tiny$\pm$0.6} & 32.9{\tiny$\pm$0.6} & 30.0{\tiny$\pm$0.3} & \cellcolor{ourshl}\textbf{48.1{\tiny$\pm$1.1}} \\
    \bottomrule
  \end{tabular}
\end{table}

\subsection{Scaling with Synthetic Data Budget}
\label{sec:scaling}

We evaluate MDA-DD with 5,000 synthetic pairs on
Flickr30K and MS-COCO to examine its effectiveness
at larger data budgets.
As shown in Table~\ref{tab:scaling}, increasing the
budget from 500 to 5,000 pairs improves mR by
5.2 and 7.4 points, respectively.
However, CovMatch remains below equally sized
real-data subsets across all six retrieval metrics
on both datasets. This suggests that increasing
the synthetic budget alone may not overcome the
limitations of feature matching along repeatedly
sampled model trajectories.
In contrast, MDA-DD incorporates supervision
from models at different training stages through
the dynamic staggered queue, while ADP and SOCM
provide complementary cross-modal matching constraints.
At 5,000 pairs, MDA-DD surpasses the real-data
baselines across all six metrics, achieving
6.1 and 3.9 higher mR points on Flickr30K and
MS-COCO, respectively.

\begin{table*}[t]
  \centering
  \caption{Scaling performance on Flickr30K and MS-COCO
  with larger synthetic-data budgets (NFNet+BERT).
  The real-data baseline uses equally sized subsets
  sampled from the original training data.
  mR is the mean of six retrieval metrics (\%).}
  \label{tab:scaling}
  \small
  \setlength{\tabcolsep}{4pt}
  \begin{tabular}{lllccccccc}
    \toprule
    Dataset & Pairs & Method
    & IR@1 & IR@5 & IR@10
    & TR@1 & TR@5 & TR@10 & mR \\
    \midrule
    \multirow{3}{*}{Flickr30K}
    & \multirow{3}{*}{5K}
    & Real Data
    & 26.6 & 57.4 & 69.6
    & 34.3 & 64.3 & 76.5 & 54.8 \\
    & & CovMatch
    & 24.0 & 53.9 & 64.9
    & 32.1 & 61.7 & 73.3 & 51.7 \\
    & & \textbf{MDA-DD}
    & \textbf{31.7} & \textbf{61.7}
    & \textbf{73.0} & \textbf{43.2}
    & \textbf{73.5} & \textbf{82.1}
    & \textbf{60.9} \\
    \midrule
    \multirow{3}{*}{MS-COCO}
    & \multirow{3}{*}{5K}
    & Real Data
    & 11.9 & 32.2 & 45.4
    & 15.3 & 38.3 & 51.0 & 32.4 \\
    & & CovMatch
    & 10.8 & 29.7 & 42.8
    & 15.1 & 36.8 & 50.4 & 30.9 \\
    & & \textbf{MDA-DD}
    & \textbf{12.9} & \textbf{34.6}
    & \textbf{47.6} & \textbf{19.3}
    & \textbf{44.7} & \textbf{58.5}
    & \textbf{36.3} \\
    \bottomrule
  \end{tabular}
  \vspace{-8pt}
\end{table*}

\begin{figure}[t]
  \centering
  \includegraphics[width=1\textwidth, trim=60 30 60 40, clip]{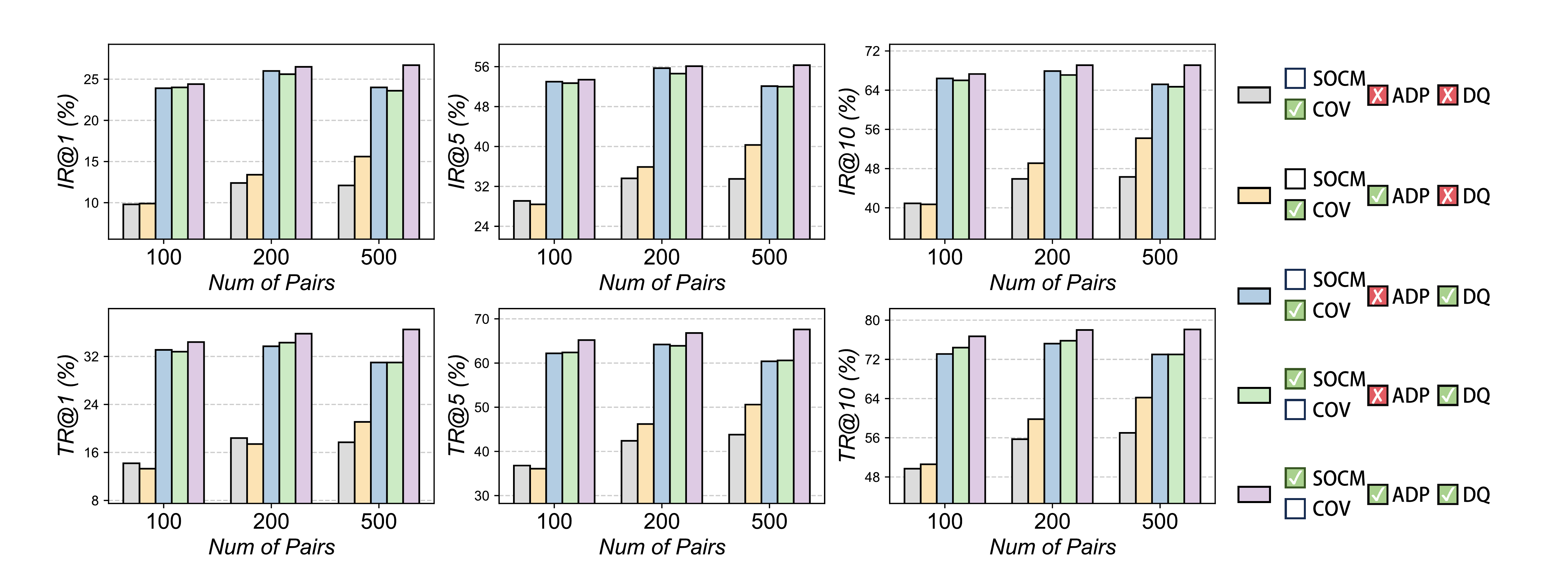}
  \caption{Component-wise ablation on Flickr30K with NFNet+BERT.
  We compare centered covariance (COV) and SOCM under different
  configurations of asymmetric dual projection (ADP) and
  dynamic model queue (DQ).}
  \label{fig:component}
  \vspace{-8pt}
\end{figure}

\begin{table*}[t]
  \centering
  \caption{Effects of asymmetric dual projection (ADP)
during distillation and downstream training on
Flickr30K with 500 synthetic pairs.
Original denotes the CovMatch projection design.
Results use the controlled ablation setting,
and mR averages the six retrieval metrics.}
  \label{tab:adp-controlled}
  \small
  \setlength{\tabcolsep}{5pt}
  \begin{tabular}{llccccccc}
    \toprule
    Distillation & Downstream
    & IR@1 & IR@5 & IR@10
    & TR@1 & TR@5 & TR@10 & mR \\
    \midrule
    Original & Original
    & 12.1 & 33.5 & 46.3
    & 17.7 & 43.8 & 57.0 & 35.1 \\
    Original & ADP
    & 15.3 & 39.1 & 52.1
    & 18.3 & 49.3 & 62.5 & 39.4 \\
    ADP & Original
    & 15.4 & 37.7 & 49.8
    & 20.8 & 45.9 & 58.2 & 38.0 \\
    ADP & ADP
    & \textbf{15.6} & \textbf{40.3} & \textbf{54.2}
    & \textbf{21.1} & \textbf{50.6}
    & \textbf{64.2} & \textbf{41.0} \\
    \bottomrule
  \end{tabular}
  \vspace{-8pt}
\end{table*}

\subsection{Ablation Study}
\label{sec:ablation}

\paragraph{Component-Wise Contribution.}
Figure~\ref{fig:component} evaluates ADP, DQ, and SOCM on
Flickr30K across three synthetic-data budgets.
DQ provides the largest retrieval gains, with ADP
yielding further improvements.
With DQ enabled and ADP disabled, replacing centered
covariance (COV) with SOCM changes mean Recall (mR)
by $+0.10$, $-0.23$, and $-0.13$ points at 100, 200,
and 500 pairs, respectively.
In a separate timing comparison with DQ and ADP fixed,
SOCM reduces moment-matching computation by 1.8\%
and full iteration time by 0.3\% at 500 pairs,
while maintaining comparable retrieval performance.

\paragraph{Distillation-Stage Effect of ADP.}
To distinguish improvements in synthetic-data quality
from downstream architectural gains, we independently
vary the projection designs during distillation and
downstream training.
As shown in Table~\ref{tab:adp-controlled}, introducing
ADP during distillation improves mR from 35.1 to 38.0
and from 39.4 to 41.0 under the two fixed downstream
architectures, respectively.
These gains across all six retrieval metrics indicate
that ADP benefits the distillation process itself,
rather than merely providing a stronger downstream model.
By introducing alignment in both visual and linguistic
spaces, ADP changes the online model training and
projected feature-mean constraints that guide
synthetic-data optimization.
Moreover, under the same ADP architecture,
MDA-DD achieves 55.7 mR compared with 41.0 for
CovMatch, showing that its performance advantage
extends beyond the projection design.

\subsection{Cross-Architecture Generalization.}
Distribution-matching methods align feature-level statistics
rather than architecture-specific parameter trajectories,
making distilled data less tied to the encoder architecture.
Following prior work, we distill with NFNet+BERT and evaluate
on alternative image encoders, including
ResNet-50~\cite{brockcharacterizing} and
RegNet~\cite{xu2022regnet}, with the text encoder held fixed.
As detailed in Table~\ref{tab:evaluation-results},
\textsc{MDA-DD} consistently outperforms the compared
baselines across both architectures.

\subsection{Computational Efficiency}
\begin{wraptable}{r}{0.48\textwidth}
  \vspace{-12pt}
  \centering
  \caption{Computational cost on MS-COCO with
100 synthetic pairs (NVIDIA RTX 5090, 32 GB).}
  \label{tab:cost_comparison}
  \begin{tabular}{lccc}
    \toprule
    \multirow{2}{*}{Method} & \multicolumn{2}{c}{Expert} & Distill \\
    \cmidrule(r){2-3} \cmidrule(l){4-4}
    & storage & time & time \\
    \midrule
    RepBlend & 16.6\,GB & 78.6\,h & 1.5\,h \\
    CovMatch & 0 & 2.0\,h & 2.8\,h \\
    \textsc{MDA-DD} & 0 & 2.0\,h & 1.3\,h \\
    \bottomrule
  \end{tabular}
  \vspace{-8pt}
\end{wraptable}

Table~\ref{tab:cost_comparison} compares the computational
costs of trajectory-based and distribution-matching methods.
RepBlend incurs substantial overhead from generating
and storing expert trajectories, whereas CovMatch and
\textsc{MDA-DD} avoid this checkpoint storage.
Notably, despite maintaining multiple online models
at different training stages, \textsc{MDA-DD} achieves
shorter distillation time than CovMatch.
Thus, broader model-state supervision need not
increase distillation cost.

\subsection{Beyond Curated Benchmarks}
\label{sec:cc3m}

We further evaluate MDA-DD on LLaVA-CC3M, using
334K image-text pairs for distillation and a disjoint
10K-pair evaluation set, with identical backbones and
optimization settings across methods.
As shown in Table~\ref{tab:cc3m}, MDA-DD outperforms
all baselines across six retrieval metrics at both
200 and 500 pairs. Notably, with only 200 pairs,
MDA-DD surpasses all 500-pair baselines, suggesting
that its multi-state supervision and cross-modal
matching design can better utilize limited synthetic
data capacity on larger web-sourced corpora.

\begin{table*}[t]
  \centering
  \caption{Cross-modal retrieval performance under alternative
  image encoder architectures on Flickr30K with 500 synthetic
  pairs. Results are reported as the mean and standard
  deviation over 5 independent runs.}
  \label{tab:evaluation-results}
  \small
  \setlength{\tabcolsep}{3.8pt}
  \begin{tabular}{ll cccccc}
    \toprule
    Backbone & Method & IR@1 & IR@5 & IR@10
      & TR@1 & TR@5 & TR@10 \\
    \midrule
    \multirow{4}{*}{ResNet+BERT}
    & LoRS
      & 3.3{\tiny$\pm$0.2}
      & 12.7{\tiny$\pm$0.3}
      & 20.4{\tiny$\pm$0.2}
      & 6.8{\tiny$\pm$0.2}
      & 19.6{\tiny$\pm$1.3}
      & 31.1{\tiny$\pm$0.3} \\
    & RepBlend
      & 4.2{\tiny$\pm$0.2}
      & 14.1{\tiny$\pm$0.2}
      & 23.6{\tiny$\pm$0.6}
      & 8.4{\tiny$\pm$0.2}
      & 23.1{\tiny$\pm$0.8}
      & 35.0{\tiny$\pm$1.3} \\
    & CovMatch
      & 10.6{\tiny$\pm$0.8}
      & 29.0{\tiny$\pm$1.3}
      & 41.0{\tiny$\pm$1.2}
      & 13.4{\tiny$\pm$1.3}
      & 34.2{\tiny$\pm$0.9}
      & 46.2{\tiny$\pm$1.4} \\
    & \textbf{Ours}
      & \textbf{14.5}{\tiny$\pm$2.2}
      & \textbf{36.4}{\tiny$\pm$4.6}
      & \textbf{49.1}{\tiny$\pm$5.2}
      & \textbf{17.8}{\tiny$\pm$2.2}
      & \textbf{42.2}{\tiny$\pm$4.5}
      & \textbf{54.1}{\tiny$\pm$5.3} \\
    \midrule
    \multirow{4}{*}{RegNet+BERT}
    & LoRS
      & 3.5{\tiny$\pm$0.1}
      & 12.6{\tiny$\pm$0.3}
      & 21.1{\tiny$\pm$0.4}
      & 6.8{\tiny$\pm$0.3}
      & 20.8{\tiny$\pm$0.3}
      & 30.2{\tiny$\pm$0.3} \\
    & RepBlend
      & 3.9{\tiny$\pm$0.2}
      & 13.9{\tiny$\pm$0.3}
      & 24.0{\tiny$\pm$0.6}
      & 7.9{\tiny$\pm$0.3}
      & 24.2{\tiny$\pm$0.3}
      & 36.2{\tiny$\pm$1.1} \\
    & CovMatch
      & 10.6{\tiny$\pm$0.3}
      & 29.5{\tiny$\pm$0.7}
      & 41.0{\tiny$\pm$0.8}
      & 13.9{\tiny$\pm$0.7}
      & 33.5{\tiny$\pm$1.0}
      & 44.8{\tiny$\pm$0.5} \\
    & \textbf{Ours}
      & \textbf{16.1}{\tiny$\pm$2.1}
      & \textbf{40.6}{\tiny$\pm$3.2}
      & \textbf{53.8}{\tiny$\pm$3.4}
      & \textbf{21.3}{\tiny$\pm$2.6}
      & \textbf{46.5}{\tiny$\pm$2.8}
      & \textbf{59.6}{\tiny$\pm$2.9} \\
    \bottomrule
  \end{tabular}
  \vspace{-8pt}
\end{table*}

\begin{table*}[t]
  \centering
  \caption{Cross-modal retrieval on LLaVA-CC3M
  with NFNet+BERT (\%).
  The full-data reference achieves
  IR@1/5/10 = 23.77/50.95/62.84 and
  TR@1/5/10 = 24.66/50.82/63.58.}
  \label{tab:cc3m}
  \small
  \setlength{\tabcolsep}{5pt}
  \begin{tabular}{llcccccc}
    \toprule
    Pairs & Method
    & IR@1 & IR@5 & IR@10
    & TR@1 & TR@5 & TR@10 \\
    \midrule
    \multirow{4}{*}{200}
    & LoRS
    & 1.56 & 5.33 & 8.77
    & 1.01 & 4.04 & 6.81 \\
    & RepBlend
    & 3.53 & 12.38 & 19.33
    & 4.39 & 13.45 & 20.18 \\
    & CovMatch
    & 2.04 & 6.95 & 12.31
    & 2.67 & 9.46 & 14.93 \\
    & \textbf{MDA-DD}
    & \textbf{7.92} & \textbf{22.29}
    & \textbf{32.13} & \textbf{10.21}
    & \textbf{28.54} & \textbf{39.62} \\
    \midrule
    \multirow{4}{*}{500}
    & LoRS
    & 1.96 & 6.72 & 10.55
    & 1.41 & 5.11 & 8.51 \\
    & RepBlend
    & 4.45 & 15.09 & 22.31
    & 5.14 & 14.89 & 22.97 \\
    & CovMatch
    & 3.00 & 10.06 & 16.07
    & 3.83 & 13.04 & 20.14 \\
    & \textbf{MDA-DD}
    & \textbf{8.88} & \textbf{24.76}
    & \textbf{35.09} & \textbf{10.23}
    & \textbf{28.99} & \textbf{40.06} \\
    \bottomrule
  \end{tabular}
  \vspace{-8pt}
\end{table*}

\section{Conclusion}
\label{sec:conclusion}
We propose MDA-DD, which extends feature matching across models at different training stages through a dynamic staggered queue. Combined with asymmetric dual projection and uncentered second-order cross-modal moment matching, MDA-DD outperforms existing methods on Flickr30K and MS-COCO across synthetic-data budgets. The distilled datasets also achieve improved retrieval performance under alternative visual encoder architectures.

\paragraph{Limitations and Future Work.}
Despite the strong performance of \textsc{MDA-DD}, current MDD frameworks, including our framework, rely heavily on CLIP-style dual-encoder architectures, which limits their direct applicability to emerging native MLLMs. The dependence on explicit visual encoders also hinders the transferability of distilled datasets to unified autoregressive generation paradigms. In future work, we plan to explore architecture-agnostic distillation strategies for native MLLMs.

\section*{Acknowledgement}
This project is supported by the Ministry of Education, Singapore, under the Academic Research Fund Tier~1 (FY2026, WBS: A-8004345-00-00),  the National Research Foundation, Singapore, and Cyber Security Agency of Singapore under its National Cybersecurity R\&D Programme and CyberSG R\&D Cyber Research Programme Office (Award: CRPO-GC1-NTU-002).

\newpage

\bibliographystyle{plain}
\bibliography{neurips_2026}

@inproceedings{brockcharacterizing,
  title={Characterizing signal propagation to close the performance gap in unnormalized ResNets},
  author={Brock, Andrew and De, Soham and Smith, Samuel L},
  booktitle={International Conference on Learning Representations}
}

@inproceedings{shin2023loss,
  title={Loss-curvature matching for dataset selection and condensation},
  author={Shin, Seungjae and Bae, Heesun and Shin, Donghyeok and Joo, Weonyoung and Moon, Il-Chul},
  booktitle={International Conference on Artificial Intelligence and Statistics},
  pages={8606--8628},
  year={2023},
  organization={PMLR}
}

@inproceedings{lee2022dataset,
  title={Dataset condensation with contrastive signals},
  author={Lee, Saehyung and Chun, Sanghyuk and Jung, Sangwon and Yun, Sangdoo and Yoon, Sungroh},
  booktitle={International Conference on Machine Learning},
  pages={12352--12364},
  year={2022},
  organization={PMLR}
}

@inproceedings{du2023minimizing,
  title={Minimizing the accumulated trajectory error to improve dataset distillation},
  author={Du, Jiawei and Jiang, Yidi and Tan, Vincent YF and Zhou, Joey Tianyi and Li, Haizhou},
  booktitle={Proceedings of the IEEE/CVF conference on computer vision and pattern recognition},
  pages={3749--3758},
  year={2023}
}

@inproceedings{du2023sequential,
  title={Sequential Subset Matching for Dataset Distillation},
  author={Du, Jiawei and Shi, Qin and Zhou, Joey Tianyi},
  booktitle={Thirty-seventh Conference on Neural Information Processing Systems}
}

@inproceedings{sajedi2023datadam,
  title={Datadam: Efficient dataset distillation with attention matching},
  author={Sajedi, Ahmad and Khaki, Samir and Amjadian, Ehsan and Liu, Lucy Z and Lawryshyn, Yuri A and Plataniotis, Konstantinos N},
  booktitle={Proceedings of the IEEE/CVF International Conference on Computer Vision},
  pages={17097--17107},
  year={2023}
}

@inproceedings{sun2024diversity,
  title={On the diversity and realism of distilled dataset: An efficient dataset distillation paradigm},
  author={Sun, Peng and Shi, Bei and Yu, Daiwei and Lin, Tao},
  booktitle={Proceedings of the IEEE/CVF conference on computer vision and pattern recognition},
  pages={9390--9399},
  year={2024}
}

@article{yin2023squeeze,
  title={Squeeze, recover and relabel: Dataset condensation at imagenet scale from a new perspective},
  author={Yin, Zeyuan and Xing, Eric and Shen, Zhiqiang},
  journal={Advances in Neural Information Processing Systems},
  volume={36},
  pages={73582--73603},
  year={2023}
}

@article{huiminunidetox,
  title={Unidetox: Universal detoxification of large language models via dataset distillation},
  author={Lu, Huimin and Isonuma, Masaru and Mori, Junichiro and Sakata, Ichiro},
  journal={arXiv preprint arXiv:2504.20500},
  year={2025}
}

@inproceedings{ding2025condensing,
  title={Condensing action segmentation datasets via generative network inversion},
  author={Ding, Guodong and Chen, Rongyu and Yao, Angela},
  booktitle={Proceedings of the Computer Vision and Pattern Recognition Conference},
  pages={17733--17742},
  year={2025}
}

@inproceedings{wang2024dancing,
  title={Dancing with still images: Video distillation via static-dynamic disentanglement},
  author={Wang, Ziyu and Xu, Yue and Lu, Cewu and Li, Yong-Lu},
  booktitle={Proceedings of the IEEE/CVF Conference on Computer Vision and Pattern Recognition},
  pages={6296--6304},
  year={2024}
}

@inproceedings{liugraph,
  title={Graph Distillation with Eigenbasis Matching},
  author={Liu, Yang and Bo, Deyu and Shi, Chuan},
  booktitle={Forty-first International Conference on Machine Learning}
}

@inproceedings{tao2024textual,
  title={Textual dataset distillation via language model embedding},
  author={Tao, Yefan and Kong, Luyang and Kan, Andrey and Callot, Laurent},
  booktitle={Findings of the Association for Computational Linguistics: EMNLP 2024},
  pages={12557--12569},
  year={2024}
}

@article{maekawa2025dataset,
  title={Dataset distillation with attention labels for fine-tuning BERT},
  author={Maekawa, Aru and Kobayashi, Naoki and Funakoshi, Kotaro and Okumura, Manabu},
  journal={Journal of Natural Language Processing},
  volume={32},
  number={1},
  pages={283--299},
  year={2025},
  publisher={The Association for Natural Language Processing}
}

@article{zhao2024video,
  title={Video set distillation: Information diversification and temporal densification},
  author={Zhao, Yinjie and Zhao, Heng and Wen, Bihan and Ong, Yew-Soon and Zhou, Joey Tianyi},
  journal={arXiv preprint arXiv:2412.00111},
  year={2024}
}

@article{li2025video,
  title={Video Dataset Condensation with Diffusion Models},
  author={Li, Zhe and Reynaud, Hadrien and Dombrowski, Mischa and Cechnicka, Sarah and Erick, Franciskus Xaverius and Kainz, Bernhard},
  year={2025}
}

@article{yang2023does,
  title={Does graph distillation see like vision dataset counterpart?},
  author={Yang, Beining and Wang, Kai and Sun, Qingyun and Ji, Cheng and Fu, Xingcheng and Tang, Hao and You, Yang and Li, Jianxin},
  journal={Advances in Neural Information Processing Systems},
  volume={36},
  pages={53201--53226},
  year={2023}
}

@inproceedings{jingraph,
  title={Graph Condensation for Graph Neural Networks},
  author={Jin, Wei and Zhao, Lingxiao and Zhang, Shichang and Liu, Yozen and Tang, Jiliang and Shah, Neil},
  booktitle={International Conference on Learning Representations}
}

@article{maekawa2025dilm,
  title={Dilm: Distilling dataset into language model for text-level dataset distillation},
  author={Maekawa, Aru and Kosugi, Satoshi and Funakoshi, Kotaro and Okumura, Manabu},
  journal={Journal of Natural Language Processing},
  volume={32},
  number={1},
  pages={252--282},
  year={2025},
  publisher={The Association for Natural Language Processing}
}

@article{yindataset,
  title={Dataset Distillation via Curriculum Data Synthesis in Large Data Era},
  author={Yin, Zeyuan and Shen, Zhiqiang},
  journal={Transactions on Machine Learning Research}
}

@article{xu2022regnet,
  title={RegNet: Self-regulated network for image classification},
  author={Xu, Jing and Pan, Yu and Pan, Xinglin and Hoi, Steven and Yi, Zhang and Xu, Zenglin},
  journal={IEEE transactions on neural networks and learning systems},
  volume={34},
  number={11},
  pages={9562--9567},
  year={2022},
  publisher={IEEE}
}

@inproceedings{welling2009herding,
  title={Herding dynamical weights to learn},
  author={Welling, Max},
  booktitle={Proceedings of the 26th annual international conference on machine learning},
  pages={1121--1128},
  year={2009}
}

@inproceedings{brock2021high,
  title={High-performance large-scale image recognition without normalization},
  author={Brock, Andy and De, Soham and Smith, Samuel L and Simonyan, Karen},
  booktitle={International conference on machine learning},
  pages={1059--1071},
  year={2021},
  organization={PMLR}
}

@book{farahani2009facility,
  title={Facility location: concepts, models, algorithms and case studies},
  author={Farahani, Reza Zanjirani and Hekmatfar, Masoud},
  year={2009},
  publisher={Springer Science \& Business Media}
}

@inproceedings{cazenavette2022dataset,
  title={Dataset distillation by matching training trajectories},
  author={Cazenavette, George and Wang, Tongzhou and Torralba, Antonio and Efros, Alexei A and Zhu, Jun-Yan},
  booktitle={Proceedings of the IEEE/CVF conference on computer vision and pattern recognition},
  pages={4750--4759},
  year={2022}
}

@inproceedings{cui2023scaling,
  title={Scaling up dataset distillation to imagenet-1k with constant memory},
  author={Cui, Justin and Wang, Ruochen and Si, Si and Hsieh, Cho-Jui},
  booktitle={International Conference on Machine Learning},
  pages={6565--6590},
  year={2023},
  organization={PMLR}
}

@inproceedings{devlin2019bert,
  title={Bert: Pre-training of deep bidirectional transformers for language understanding},
  author={Devlin, Jacob and Chang, Ming-Wei and Lee, Kenton and Toutanova, Kristina},
  booktitle={Proceedings of the 2019 conference of the North American chapter of the association for computational linguistics: human language technologies, volume 1 (long and short papers)},
  pages={4171--4186},
  year={2019}
}

@inproceedings{dosovitskiyimage,
  title={An Image is Worth 16x16 Words: Transformers for Image Recognition at Scale},
  author={Dosovitskiy, Alexey and Beyer, Lucas and Kolesnikov, Alexander and Weissenborn, Dirk and Zhai, Xiaohua and Unterthiner, Thomas and Dehghani, Mostafa and Minderer, Matthias and Heigold, Georg and Gelly, Sylvain and others},
  booktitle={International Conference on Learning Representations}
}

@inproceedings{guomultimodal,
  title={Multimodal Dataset Distillation via Phased Teacher Models},
  author={Guo, Shengbin and Zhao, Hang and Yang, Senqiao and Jiang, Chenyang and Cheng, Yuhang and Peng, Xiangru and Shao, Rui and Tian, Zhuotao},
  booktitle={The Fourteenth International Conference on Learning Representations}
}

@inproceedings{karpathy2015deep,
  title={Deep visual-semantic alignments for generating image descriptions},
  author={Karpathy, Andrej and Fei-Fei, Li},
  booktitle={Proceedings of the IEEE conference on computer vision and pattern recognition},
  pages={3128--3137},
  year={2015}
}

@inproceedings{leecovmatch,
  title={CovMatch: Cross-Covariance Guided Multimodal Dataset Distillation with Trainable Text Encoder},
  author={Lee, Yongmin and Chung, Hye Won},
  booktitle={The Thirty-ninth Annual Conference on Neural Information Processing Systems}
}

@inproceedings{lin2014microsoft,
  title={Microsoft coco: Common objects in context},
  author={Lin, Tsung-Yi and Maire, Michael and Belongie, Serge and Hays, James and Perona, Pietro and Ramanan, Deva and Doll{\'a}r, Piotr and Zitnick, C Lawrence},
  booktitle={European conference on computer vision},
  pages={740--755},
  year={2014},
  organization={Springer}
}

@article{oord2018representation,
  title={Representation learning with contrastive predictive coding},
  author={Oord, Aaron van den and Li, Yazhe and Vinyals, Oriol},
  journal={arXiv preprint arXiv:1807.03748},
  year={2018}
}

@inproceedings{plummer2015flickr30k,
  title={Flickr30k entities: Collecting region-to-phrase correspondences for richer image-to-sentence models},
  author={Plummer, Bryan A and Wang, Liwei and Cervantes, Chris M and Caicedo, Juan C and Hockenmaier, Julia and Lazebnik, Svetlana},
  booktitle={Proceedings of the IEEE international conference on computer vision},
  pages={2641--2649},
  year={2015}
}

@inproceedings{radford2021learning,
  title={Learning transferable visual models from natural language supervision},
  author={Radford, Alec and Kim, Jong Wook and Hallacy, Chris and Ramesh, Aditya and Goh, Gabriel and Agarwal, Sandhini and Sastry, Girish and Askell, Amanda and Mishkin, Pamela and Clark, Jack and others},
  booktitle={International conference on machine learning},
  pages={8748--8763},
  year={2021},
  organization={PmLR}
}

@inproceedings{tonevaempirical,
  title={An Empirical Study of Example Forgetting during Deep Neural Network Learning},
  author={Toneva, Mariya and Sordoni, Alessandro and des Combes, Remi Tachet and Trischler, Adam and Bengio, Yoshua and Gordon, Geoffrey J},
  booktitle={International Conference on Learning Representations}
}

@article{tschannen2025siglip,
  title={Siglip 2: Multilingual vision-language encoders with improved semantic understanding, localization, and dense features},
  author={Tschannen, Michael and Gritsenko, Alexey and Wang, Xiao and Naeem, Muhammad Ferjad and Alabdulmohsin, Ibrahim and Parthasarathy, Nikhil and Evans, Talfan and Beyer, Lucas and Xia, Ye and Mustafa, Basil and others},
  journal={arXiv preprint arXiv:2502.14786},
  year={2025}
}

@article{wang2018dataset,
  title={Dataset Distillation},
  author={Wang, Tongzhou and Zhu, Jun-Yan and Torralba, Antonio and Efros, Alexei A}
}

@article{wuvision,
  title={Vision-Language Dataset Distillation},
  author={Wu, Xindi and Zhang, Byron and Deng, Zhiwei and Russakovsky, Olga},
  journal={Transactions on Machine Learning Research}
}

@inproceedings{xu2024low,
  title={Low-Rank Similarity Mining for Multimodal Dataset Distillation},
  author={Xu, Yue and Lin, Zhilin and Qiu, Yusong and Lu, Cewu and Li, Yong-Lu},
  booktitle={Forty-first International Conference on Machine Learning}
}

@inproceedings{zhai2023sigmoid,
  title={Sigmoid loss for language image pre-training},
  author={Zhai, Xiaohua and Mustafa, Basil and Kolesnikov, Alexander and Beyer, Lucas},
  booktitle={Proceedings of the IEEE/CVF international conference on computer vision},
  pages={11975--11986},
  year={2023}
}

@inproceedings{zhangbeyond,
  title={Beyond Modality Collapse: Representation Blending for Multimodal Dataset Distillation},
  author={Zhang, Xin and Zhang, Ziruo and Du, Jiawei and Liu, Zuozhu and Zhou, Joey Tianyi},
  booktitle={The Thirty-ninth Annual Conference on Neural Information Processing Systems}
}

@inproceedings{zhang2024navigating,
  title={Navigating Complexity: Toward Lossless Graph Condensation via Expanding Window Matching},
  author={Zhang, Yuchen and Zhang, Tianle and Wang, Kai and Guo, Ziyao and Liang, Yuxuan and Bresson, Xavier and Jin, Wei and You, Yang},
  booktitle={Forty-first International Conference on Machine Learning}
}

@inproceedings{jia2021scaling,
  title={Scaling up visual and vision-language representation learning with noisy text supervision},
  author={Jia, Chao and Yang, Yinfei and Xia, Ye and Chen, Yi-Ting and Parekh, Zarana and Pham, Hieu and Le, Quoc and Sung, Yun-Hsuan and Li, Zhen and Duerig, Tom},
  booktitle={International conference on machine learning},
  pages={4904--4916},
  year={2021},
  organization={PMLR}
}

@inproceedings{zhao2021dc,
  title={Dataset Condensation with Gradient Matching},
  author={Zhao, Bo and Mopuri, Konda Reddy and Bilen, Hakan},
  booktitle={International Conference on Learning Representations}
}

@inproceedings{zhao2021dsa,
  title={Dataset condensation with differentiable siamese augmentation},
  author={Zhao, Bo and Bilen, Hakan},
  booktitle={International Conference on Machine Learning},
  pages={12674--12685},
  year={2021},
  organization={PMLR}
}

@inproceedings{zhao2023dm,
  title={Dataset condensation with distribution matching},
  author={Zhao, Bo and Bilen, Hakan},
  booktitle={Proceedings of the IEEE/CVF winter conference on applications of computer vision},
  pages={6514--6523},
  year={2023}
}

@inproceedings{wang2022cafe,
  title={Cafe: Learning to condense dataset by aligning features},
  author={Wang, Kai and Zhao, Bo and Peng, Xiangyu and Zhu, Zheng and Yang, Shuo and Wang, Shuo and Huang, Guan and Bilen, Hakan and Wang, Xinchao and You, Yang},
  booktitle={Proceedings of the IEEE/CVF Conference on Computer Vision and Pattern Recognition},
  pages={12196--12205},
  year={2022}
}

@inproceedings{zhao2023idm,
  title={Improved Distribution Matching for Dataset Condensation},
  author={Zhao, Ganlong and Li, Guanbin and Qin, Yipeng and Yu, Yizhou},
  booktitle={2023 IEEE/CVF Conference on Computer Vision and Pattern Recognition (CVPR)},
  pages={7856--7865},
  year={2023},
  organization={IEEE Computer Society}
}

@article{van2008tsne,
  title={Visualizing data using t-SNE.},
  author={Van der Maaten, Laurens and Hinton, Geoffrey},
  journal={Journal of machine learning research},
  volume={9},
  number={11},
  year={2008}
}

@article{kobak2019art,
  title={The art of using t-SNE for single-cell transcriptomics},
  author={Kobak, Dmitry and Berens, Philipp},
  journal={Nature communications},
  volume={10},
  number={1},
  pages={5416},
  year={2019},
  publisher={Nature Publishing Group UK London}
}

@article{yu2023dataset,
  title={Dataset distillation: A comprehensive review},
  author={Yu, Ruonan and Liu, Songhua and Wang, Xinchao},
  journal={IEEE transactions on pattern analysis and machine intelligence},
  volume={46},
  number={1},
  pages={150--170},
  year={2023},
  publisher={IEEE}
}

@inproceedings{bo2026efficient,
  title={Efficient Multi-modal Dataset Distillation via Analytic Parameter Matching},
  author={Bo, Deyu and Wang, Xinchao},
   booktitle={International Conference on Machine Learning},
  year={2026}
}

@inproceedings{yu2024teddy,
  title={Teddy: Efficient large-scale dataset distillation via taylor-approximated matching},
  author={Yu, Ruonan and Liu, Songhua and Ye, Jingwen and Wang, Xinchao},
  booktitle={European Conference on Computer Vision},
  pages={1--17},
  year={2024},
  organization={Springer}
}

@inproceedings{2026RuonanFlashDD,
  title={Flash-DD: An Ultra Parameter-Efficient Approach to Dataset Distillation},
  author={Yu, Ruonan and Liu, Songhua and Wang, Xinchao},
  booktitle={European Conference on Computer Vision},
  year={2026},
}

@inproceedings{2026LingaoDD,
  title={Unifying Dataset Pruning and Distillation for Efficient Large-scale Compression},
  author={Xiao, Lingao and  Liu, Songhua and He, Yang and Wang, Xinchao},
  booktitle={International Conference on Machine Learning},
  year={2026},
}

@inproceedings{2026DeyuCVPR,
  title={Beyond Soft Label: Dataset Distillation via Orthogonal Gradient Matching},
  author={Bo, Deyu and Wang, Xinchao},
  booktitle={IEEE/CVF Conference on Computer Vision and Pattern Recognition},
  year={2026},
}

@inproceedings{2026OLetianICLR,
  title={{CoDA: From Text-to-Image Diffusion Models to Training-Free Dataset Distillation}},
  author={Zhou, Letian and  Liu, Songhua and Wang, Xinchao},
  booktitle={International Conference on Learning Representations},
  year={2026}
}

@inproceedings{2026ODeyuICLR,
  title={{Understanding Dataset Distillation via Spectral Filtering}},
  author={Bo, Deyu and  Liu, Songhua and Wang, Xinchao},
  booktitle={International Conference on Learning Representations},
  year={2026}
}

@inproceedings{2026Zigengdparallel,
  title={{dParallel: Learnable Parallel Decoding for dLLMs}},
  author={Chen, Zigeng and  Fang, Gongfan and  Ma, Xinyin and  Yu, Ruonan and  Wang, Xinchao},
  booktitle={International Conference on Learning Representations},
  year={2026}
}

@inproceedings{2026OZigengDMax,
  title={{DMax: Aggressive Parallel Decoding for dLLMs}},
  author={Chen, Zigeng and  Fang, Gongfan and  Ma, Xinyin and  Yu, Ruonan and  Wang, Xinchao},
  booktitle={Neural Information Processing Systems},
  year={2026}
}

\newpage

\appendix

\section{More Related Work}
\label{app:related_work}

\subsection{Vision-Language Pre-Training}
\label{app:vlp}

Our work distills the training data of dual-encoder vision-language
models, a paradigm popularized by
CLIP~\cite{radford2021learning} and ALIGN~\cite{jia2021scaling}.
In this architecture, an image encoder and a text encoder independently
produce embeddings that are aligned via the InfoNCE
objective~\cite{oord2018representation} in a shared space.
SigLIP~\cite{zhai2023sigmoid} replaces the softmax-based contrastive
loss with a per-pair sigmoid loss, improving batch efficiency.
All existing multimodal dataset distillation
methods~\cite{wuvision,xu2024low,zhangbeyond,leecovmatch,guomultimodal}
operate under this dual-encoder setting. 

\subsection{Dataset Distillation}
\label{app:unimodal_dd}

Dataset distillation synthesizes a compact training set whose learning signal approximates that of the full dataset~\cite{yu2023dataset}.
Early meta-learning formulations~\cite{wang2018dataset} optimize
synthetic pixels through bi-level optimization but are computationally
prohibitive.

Gradient matching (DC~\cite{zhao2021dc}, DCC~\cite{lee2022dataset}, DSA~\cite{zhao2021dsa}, LCMat~\cite{shin2023loss})
sidesteps the inner loop by aligning single-step gradients, while
trajectory matching (MTT~\cite{cazenavette2022dataset}, TESLA~\cite{cui2023scaling}, FTD~\cite{du2023minimizing}, SeqMatch~\cite{du2023sequential}) extends this to multi-step parameter
trajectories at the cost of storing dense expert checkpoints.
Teddy~\cite{yu2024teddy} further improves distillation efficiency through Taylor-approximated matching and a pre-cached pool of weak models.
Distribution matching (DM~\cite{zhao2023dm}, DataDAM~\cite{sajedi2023datadam}, CAFE~\cite{wang2022cafe}, IDM~\cite{zhao2023idm}, RDED~\cite{sun2024diversity}, SRe$^2$L~\cite{yin2023squeeze}, CDA~\cite{yindataset}) takes a different route, directly minimizing
the discrepancy between real and synthetic feature distributions. Beyond image-centric settings, dataset distillation has also been studied for text \cite{huiminunidetox,maekawa2025dilm}, video~\cite{wang2024dancing,ding2025condensing}, and graph data~\cite{liugraph,zhang2024navigating}, where recent studies synthesize compact data by exploiting domain-specific structures, including linguistic generation~\cite{tao2024textual,maekawa2025dataset,2026Zigengdparallel,2026OZigengDMax}, temporal dynamics~\cite{zhao2024video,li2025video}, and graph spectra~\cite{yang2023does,jingraph}. These advances indicate that dataset distillation is evolving from a vision-oriented technique into a general data compression framework, further motivating its extension to multimodal learning. Among existing multimodal efforts, MTT-VL~\cite{wuvision} adapts trajectory matching to vision-language data; by contrast, distribution matching provides a more favorable efficiency--quality trade-off and serves as the methodological basis of CovMatch~\cite{leecovmatch} and our method. We follow this paradigm but address challenges specific to multimodal settings: cross-modal alignment between heterogeneous feature spaces, dimensional asymmetry between encoders, and the lack of discrete class labels that can serve as alignment anchors.

\subsection{Trajectory-Based Multimodal Distillation}
\label{app:trajectory}

Trajectory-based methods extend MTT to vision-language models by
matching expert parameter trajectories across both encoder branches.
Given a real dataset $\mathcal{T}$, an expert trajectory
$\tau^{*}=\{\Theta^{*}_{t}\}_{t=0}^{T}$ is pre-computed by training
a vision-language model with the symmetric InfoNCE loss
(Eq.~\ref{eq:infonce}), where
$\Theta^{*}_{t}=(\Theta^{*}_{\mathrm{img},t},\,
\Theta^{*}_{\mathrm{txt},t})$. A student model initialized from an
expert checkpoint $\hat{\Theta}_{s}=\Theta^{*}_{s}$ is then trained on the
synthetic data $\mathcal{S}$ for $R'$ inner steps, and the synthetic
data is updated by minimizing the bi-modal trajectory discrepancy:
\begin{equation}\label{eq:bimtt_app}
  \mathcal{L}_{\text{bi-traj}}
  \;=\;
  \frac{\|\hat{\Theta}_{\mathrm{img},s+R'}-\Theta^{*}_{\mathrm{img},s+R}\|^{2}}
       {\|\Theta^{*}_{\mathrm{img},s}-\Theta^{*}_{\mathrm{img},s+R}\|^{2}}
  \;+\;
  \frac{\|\hat{\Theta}_{\mathrm{txt},s+R'}-\Theta^{*}_{\mathrm{txt},s+R}\|^{2}}
       {\|\Theta^{*}_{\mathrm{txt},s}-\Theta^{*}_{\mathrm{txt},s+R}\|^{2}}\,,
\end{equation}
where $R$ is the expert matching horizon and $R'$ is the
student unrolling horizon. Representative methods include MTT-VL,
which provides the first baseline; LoRS~\cite{xu2024low}, which jointly optimizes a soft-label similarity matrix
alongside the synthetic image-text pairs and applies low-rank decomposition
to this matrix to reduce computational cost; and
RepBlend~\cite{zhangbeyond}, which freezes both encoders and blends real and synthetic representations. Despite these refinements, all trajectory-based
methods require a full pre-training run to produce the expert
trajectory, which is precisely the cost that distillation aims to avoid. 
As shown in Table~\ref{tab:cost_comparison}, RepBlend incurs 16.6~GB of
expert storage and 78.6~hours of expert training, compared with zero
additional storage and 2.0~hours for distribution-based methods
including ours. More recently, APM~\cite{bo2026efficient} proposed an efficient multimodal dataset distillation framework that matches analytically estimated projection parameters, reducing the computational and storage overhead associated with conventional trajectory-based methods.
\section{t-SNE Visualization}
\label{app:tsne}

This section describes the procedure used to generate the t-SNE
visualization in Figure~\ref{fig:TSNE}~\cite{van2008tsne}.

\paragraph{Feature Extraction.}
We instantiate a CLIPModel (NFNet image encoder + BERT text encoder)
with pre-trained weights under a random seed distinct from those used
during distillation; all parameters are frozen. For each method (Real
Data, CovMatch, \textsc{MDA-DD}), we extract image embeddings
$\mathbf{e}_v^i = f_v(\tilde{\mathbf{x}}_v^i;\theta_v)
\in \mathbb{R}^{d_v}$ without projection or normalization, and text
embeddings
$\mathbf{z}_l^i = \mathbf{W}_l^{\uparrow} \cdot
f_l(\tilde{\mathbf{x}}_l^i;\theta_l) \in \mathbb{R}^{d_v}$,
projected into the visual space for direct comparison.

\paragraph{Joint t-SNE Reduction.}
All image and text embeddings from the three configurations are
concatenated into a single matrix
$\mathbf{X} \in \mathbb{R}^{6N \times d_v}$ (2 modalities $\times$
3 methods $\times$ $N$ pairs) and jointly reduced by a single t-SNE
instance to $\mathbf{X}_{\mathrm{2D}} \in \mathbb{R}^{6N \times 2}$.
Joint reduction is essential because all embeddings are extracted by
the same model and reside in the same high-dimensional space;
reducing them simultaneously preserves their relative proximity. We
set the perplexity to 50 and use PCA initialization~\cite{kobak2019art}.

\paragraph{Observations.}
As shown in Figure~\ref{fig:TSNE}(a), real-data embeddings form two well-spread, spatially interleaved
clusters with strong cross-modal alignment. CovMatch exhibits two
pathologies: a subset of text embeddings collapses into an isolated
region far from both main clusters, consistent with the
initialization overfitting analyzed in
Section~\ref{sec:mda}. The embeddings produced by \textsc{MDA-DD}
distribute similarly to real data, with both modalities well-spread
and interleaved, confirming that initialization overfitting is
effectively alleviated.
\section{Proofs and Derivations}
\label{app:proofs}

\subsection{Matrix Form Equivalence of SOCM (Eq.~\ref{eq:socm_def})}
\label{app:proof_matrix}

Let $\mathbf{E}_v \in \mathbb{R}^{n \times d_v}$ and
$\mathbf{E}_l \in \mathbb{R}^{n \times d_l}$ be obtained by stacking
$\mathbf{e}_v^i$ and $\mathbf{e}_l^i$ row-wise. Then
$\frac{1}{n}\sum_{i=1}^{n} \mathbf{e}_v^i(\mathbf{e}_l^i)^\top
= \frac{1}{n}\mathbf{E}_v^\top\mathbf{E}_l$.

The $(p,q)$-th entry of $\mathbf{E}_v^\top\mathbf{E}_l$ is
$\sum_{i=1}^{n}[\mathbf{e}_v^i]_p[\mathbf{e}_l^i]_q$, which equals
the $(p,q)$-th entry of
$\sum_{i=1}^{n}\mathbf{e}_v^i(\mathbf{e}_l^i)^\top$. Since this
holds for all $(p,q)$, dividing by $n$ yields the result.

\subsection{SOCM Decomposition (Eq.~\ref{eq:socm_decomp})}
\label{app:proof_decomp}

Let $\boldsymbol{\mu}_v = \frac{1}{n}\sum_i \mathbf{e}_v^i$,
$\boldsymbol{\mu}_l = \frac{1}{n}\sum_i \mathbf{e}_l^i$, and
$\mathbf{C} = \frac{1}{n}\sum_i (\mathbf{e}_v^i - \boldsymbol{\mu}_v)
(\mathbf{e}_l^i - \boldsymbol{\mu}_l)^\top$. Then
$\mathbf{M}_\theta(\mathcal{D}) = \mathbf{C} + \boldsymbol{\mu}_v
\boldsymbol{\mu}_l^\top$.

Substituting $\mathbf{e}_v^i = \bar{\mathbf{e}}_v^i +
\boldsymbol{\mu}_v$ and $\mathbf{e}_l^i = \bar{\mathbf{e}}_l^i +
\boldsymbol{\mu}_l$ into the SOCM definition and expanding:
\begin{align}
  \mathbf{M}_\theta(\mathcal{D})
  &= \frac{1}{n}\sum_{i=1}^{n}
     (\bar{\mathbf{e}}_v^i + \boldsymbol{\mu}_v)
     (\bar{\mathbf{e}}_l^i + \boldsymbol{\mu}_l)^\top \notag \\
  &= \underbrace{\frac{1}{n}\sum_i \bar{\mathbf{e}}_v^i
     (\bar{\mathbf{e}}_l^i)^\top}_{\mathbf{C}}
   + \underbrace{\frac{1}{n}\Bigl(\sum_i \bar{\mathbf{e}}_v^i\Bigr)
     \boldsymbol{\mu}_l^\top}_{=\,\mathbf{0}}
   + \underbrace{\boldsymbol{\mu}_v \frac{1}{n}
     \Bigl(\sum_i \bar{\mathbf{e}}_l^i\Bigr)^\top}_{=\,\mathbf{0}}
   + \boldsymbol{\mu}_v\boldsymbol{\mu}_l^\top.
\end{align}
The cross terms vanish because
$\sum_i \bar{\mathbf{e}}_v^i = \sum_i \bar{\mathbf{e}}_l^i =
\mathbf{0}$ by definition of centering.

This decomposition shows that SOCM incorporates both the centered cross-modal covariance and the mean outer product within a single matrix. Matching the raw moment therefore preserves their combined contribution without requiring explicit mean-centering.

\subsection{Approximation Analysis of the Staggered
Model Queue
(Eq.~\ref{eq:dist_obj}$\to$Eq.~\ref{eq:particle_obj})}
\label{app:proof_particle}

We analyze how the staggered queue approximates a
model-state matching objective. The analysis first
considers a fixed synthetic dataset and then
distinguishes ideal stratified sampling from the
deterministic training ages used by the actual queue.

\paragraph{Loss and boundedness.}
For a fixed synthetic dataset $\mathcal{S}$, define
the matching loss under model state $\theta$ as
\begin{equation}
\begin{aligned}
g_{\mathcal{S}}(\theta)
={}&
\bigl\|\mathbf{M}_\theta(\mathcal{S})
-\rho\mathbf{M}_\theta(\mathcal{T})\bigr\|_F^2\\
&+\lambda\Bigl(
\bigl\|\bar{\mathbf z}_{v,\theta}^{\,\mathcal S}
-\bar{\mathbf z}_{v,\theta}^{\,\mathcal T}
\bigr\|_2^2
+
\bigl\|\bar{\mathbf z}_{l,\theta}^{\,\mathcal S}
-\bar{\mathbf z}_{l,\theta}^{\,\mathcal T}
\bigr\|_2^2
\Bigr).
\end{aligned}
\end{equation}

Assume that, over the model states under
consideration, encoder features satisfy
$\|\mathbf e_v\|_2\leq R_v$ and
$\|\mathbf e_l\|_2\leq R_l$, while projected
features satisfy
$\|\mathbf z_v\|_2\leq Z_v$ and
$\|\mathbf z_l\|_2\leq Z_l$.
For $\lambda\geq 0$, these bounds imply
\begin{equation}
0\leq g_{\mathcal S}(\theta)\leq B,
\end{equation}
where one valid upper bound is
\begin{equation}
B=(1+|\rho|)^2R_v^2R_l^2
+4\lambda(Z_v^2+Z_l^2).
\end{equation}
This boundedness condition is an explicit
assumption of the analysis, rather than a
consequence of network architecture alone.

\paragraph{Training-stage distribution.}
Let $\Theta_a$ denote the random model state
after $a$ real-data training updates, starting
from the prescribed pretrained encoder weights.
The randomness includes independently initialized
projection heads and stochastic training updates.
Write $P_a$ for the distribution of $\Theta_a$ and
\begin{equation}
m(a)=\mathbb{E}_{\theta\sim P_a}
[g_{\mathcal S}(\theta)].
\end{equation}

For integer training ages
$a\in\{0,\ldots,T_{\max}-1\}$, consider the
stage-averaged objective
\begin{equation}
\mu_{\mathrm{stage}}
=\frac{1}{T_{\max}}
\sum_{a=0}^{T_{\max}-1}m(a).
\end{equation}
This corresponds to a model-state distribution
obtained by sampling the training age uniformly
and then drawing a model state from $P_a$.
It is a specific stage-based distribution and
need not coincide with the general target
distribution $P(\Theta)$ in Eq.~\ref{eq:dist_obj}.

\subsubsection{I.i.d.\ baseline}

Suppose $\theta_1,\ldots,\theta_K$ are drawn
independently from the stage-based distribution.
Define
\begin{equation}
\bar g_K^{\mathrm{MC}}
=\frac{1}{K}\sum_{k=1}^K
g_{\mathcal S}(\theta_k).
\end{equation}
Then
$\mathbb{E}[\bar g_K^{\mathrm{MC}}]
=\mu_{\mathrm{stage}}$.
Since $g_{\mathcal S}(\theta_k)\in[0,B]$,
Popoviciu's inequality gives
\begin{equation}
\mathrm{Var}[\bar g_K^{\mathrm{MC}}]
\leq\frac{B^2}{4K}.
\end{equation}
Consequently, by Cauchy--Schwarz,
\begin{equation}
\mathbb{E}\bigl[
|\bar g_K^{\mathrm{MC}}
-\mu_{\mathrm{stage}}|
\bigr]
\leq\frac{B}{2\sqrt K}.
\end{equation}
Hoeffding's inequality further yields
\begin{equation}
\Pr\bigl(
|\bar g_K^{\mathrm{MC}}
-\mu_{\mathrm{stage}}|
\geq\epsilon
\bigr)
\leq
2\exp\left(
-\frac{2K\epsilon^2}{B^2}
\right).
\end{equation}

\subsubsection{Ideal stratified sampling}

Assume that $K$ divides $T_{\max}$ and
let $\Delta t=T_{\max}/K$.
Partition the training ages into strata
\begin{equation}
I_k=
\{(k-1)\Delta t,\ldots,k\Delta t-1\}.
\end{equation}
Let $H$ denote the stratum index of a
uniformly sampled training age.
Each stratum has probability
$\Pr(H=k)=1/K$.

Consider an ideal stratified estimator that
independently draws one model state from the
conditional distribution associated with each
stratum. Let
\begin{equation}
\mu_k=\mathbb{E}[g_{\mathcal S}(\theta)\mid H=k],
\qquad
\sigma_k^2=
\mathrm{Var}[g_{\mathcal S}(\theta)\mid H=k].
\end{equation}
The estimator
\begin{equation}
\bar g_K^{\mathrm{strat}}
=\frac{1}{K}\sum_{k=1}^K
g_{\mathcal S}(\theta_k)
\end{equation}
is unbiased for $\mu_{\mathrm{stage}}$.
Because the draws are independent across
strata,
\begin{equation}
\mathrm{Var}[
\bar g_K^{\mathrm{strat}}]
=\frac{1}{K^2}
\sum_{k=1}^K\sigma_k^2.
\end{equation}

By the law of total variance,
\begin{equation}
\begin{aligned}
\mathrm{Var}[g_{\mathcal S}(\theta)]
={}&
\frac{1}{K}\sum_{k=1}^K\sigma_k^2\\
&+\frac{1}{K}\sum_{k=1}^K
(\mu_k-\mu_{\mathrm{stage}})^2.
\end{aligned}
\end{equation}
The between-stratum term is non-negative.
Therefore,
\begin{equation}\label{eq:strat_var}
\mathrm{Var}[
\bar g_K^{\mathrm{strat}}]
\leq
\frac{
\mathrm{Var}[g_{\mathcal S}(\theta)]
}{K}
=
\mathrm{Var}[
\bar g_K^{\mathrm{MC}}].
\end{equation}
The inequality is strict when the stratum
means differ. Thus, ideal stratification
reduces sampling variance by accounting
for variation across training stages.

\subsubsection{Approximation by staggered ages}

The actual queue differs from ideal
stratified sampling because its model ages
are determined by the replacement schedule.
At steady state, Eq.~\ref{eq:particle_obj}
uses one model from each training-age stratum.
For a fixed replacement-cycle offset
$r\in\{0,\ldots,\Delta t-1\}$, the ages are
\begin{equation}
a_k(r)=(k-1)\Delta t+r,
\qquad k=1,\ldots,K.
\end{equation}
Let $\theta_k(r)$ denote the model state
at age $a_k(r)$, and define
\begin{equation}
\widehat\mu_K(r)=
\frac{1}{K}\sum_{k=1}^K
g_{\mathcal S}(\theta_k(r)).
\end{equation}

For a fixed $\mathcal S$, suppose the
model trajectories use independent training
randomness, so that the states
$\theta_k(r)$ are independent across
models. They need not be identically
distributed because their ages differ.
The estimator has expectation
\begin{equation}
\mu_r=
\mathbb{E}[\widehat\mu_K(r)]
=\frac{1}{K}\sum_{k=1}^K m(a_k(r)).
\end{equation}

Assume that the expected matching loss
varies smoothly across training ages:
\begin{equation}
|m(a)-m(b)|\leq L_a|a-b|
\end{equation}
for some constant $L_a\geq 0$.
The difference between the staggered-age
expectation and the stage-averaged
objective satisfies
\begin{equation}
\begin{aligned}
|\mu_r-\mu_{\mathrm{stage}}|
&\leq
\frac{L_a}{K\Delta t}
\sum_{k=1}^K
\sum_{a\in I_k}|a_k(r)-a|\\
&\leq
\frac{L_a(\Delta t-1)}{2}.
\end{aligned}
\end{equation}

Since the model states are independent
under the stated assumption and their
losses lie in $[0,B]$, Hoeffding's
inequality for independent,
not necessarily identically distributed,
random variables gives
\begin{equation}
\Pr\left(
|\widehat\mu_K(r)-\mu_r|
\geq\epsilon
\right)
\leq
2\exp\left(
-\frac{2K\epsilon^2}{B^2}
\right).
\end{equation}
Combining this concentration bound with
the stage-discretization error yields
\begin{equation}
\begin{aligned}
\mathbb{E}\bigl[
|\widehat\mu_K(r)
-\mu_{\mathrm{stage}}|
\bigr]
\leq
\frac{B}{2\sqrt K}
+\frac{L_a(\Delta t-1)}{2}.
\end{aligned}
\end{equation}

The first term represents stochastic
variation across independently evolving
models, while the second captures the
difference between the staggered ages
and the stage-averaged objective.
Moreover, averaging the expected
staggered estimator over a complete
replacement cycle gives
\begin{equation}
\frac{1}{\Delta t}
\sum_{r=0}^{\Delta t-1}\mu_r
=\mu_{\mathrm{stage}}.
\end{equation}
Thus, the staggered schedule covers all
training ages across a complete cycle,
although its expectation at any single
offset need not equal the stage average.

For the more general target distribution
$P(\Theta)$ in Eq.~\ref{eq:dist_obj},
define the distributional mismatch
\begin{equation}
\delta_P(\mathcal S)=
\left|
\mu_{\mathrm{stage}}
-
\mathbb{E}_{\theta\sim P(\Theta)}
[g_{\mathcal S}(\theta)]
\right|.
\end{equation}
The corresponding expected approximation
error is bounded by
\begin{equation}
\begin{aligned}
&\mathbb{E}\left[
\left|
\widehat\mu_K(r)-
\mathbb{E}_{\theta\sim P(\Theta)}
[g_{\mathcal S}(\theta)]
\right|
\right]\\
&\qquad\leq
\frac{B}{2\sqrt K}
+\frac{L_a(\Delta t-1)}{2}
+\delta_P(\mathcal S).
\end{aligned}
\end{equation}
The mismatch term vanishes when the
target distribution is the
stage-based distribution defined above.

\paragraph{Dependence and scope.}
The preceding concentration bounds hold
for a fixed synthetic dataset and
independently evolving model trajectories.
They characterize model-state objective
approximation, not convergence of the
joint optimization algorithm.

During actual distillation, the synthetic
dataset is updated using the active queue.
Consequently, the losses evaluated at the
adaptively optimized synthetic dataset
may be statistically dependent, even
when the underlying model trajectories
use independent randomness.

More generally, for dependent losses
$Y_k=g_{\mathcal S}(\theta_k)$,
the estimator variance satisfies
\begin{equation}
\begin{aligned}
\mathrm{Var}\left[
\frac{1}{K}\sum_{k=1}^K Y_k
\right]
={}&
\frac{1}{K^2}
\sum_{k=1}^K\mathrm{Var}[Y_k]\\
&+\frac{2}{K^2}
\sum_{i<j}\mathrm{Cov}(Y_i,Y_j).
\end{aligned}
\end{equation}
Thus, a comparable variance rate under
dependence requires additional control
of the cross-model covariance terms.
Such control is not implied solely by
staggered replacement or weight decay.

The analysis isolates the benefit of
training-stage coverage under explicit
sampling and regularity assumptions.
The empirical ablations in
Section~4.4 and Appendix~F further
evaluate the effectiveness of the
staggered queue in the actual
distillation procedure.

\section{Implementation of Asymmetric Dual Projection}
\label{app:implementation}

\begin{algorithm}[h]
\caption{Forward Pass with Asymmetric Dual Projection}
\label{alg:forward}
\begin{algorithmic}[1]
\REQUIRE Image batch $\{\mathbf{x}_v^i\}_{i=1}^B$,
text batch $\{(\mathbf{x}_l^i,\mathbf{m}^i)\}_{i=1}^B$,
learnable logit-scale parameter $\ell$
\ENSURE Contrastive loss $\mathcal{L}$
\STATE $\mathbf{e}_v^i \leftarrow
f_v(\mathbf{x}_v^i;\theta_v)
\in\mathbb{R}^{d_v},\quad\forall\,i$
\STATE $\mathbf{e}_l^i \leftarrow
f_l(\mathbf{x}_l^i,\mathbf{m}^i;\theta_l)
\in\mathbb{R}^{d_l},\quad\forall\,i$
\STATE $\mathbf{z}_l^i \leftarrow
\mathbf{W}_l^{\uparrow}\mathbf{e}_l^i
\in\mathbb{R}^{d_v}$
\COMMENT{text $\to$ visual space}
\STATE $\mathbf{z}_v^i \leftarrow
\mathbf{W}_v^{\downarrow}\mathbf{e}_v^i
\in\mathbb{R}^{d_l}$
\COMMENT{image $\to$ linguistic space}
\STATE $\hat{\mathbf{e}}_v^i \leftarrow
\mathbf{e}_v^i/\|\mathbf{e}_v^i\|_2,\quad
\hat{\mathbf{z}}_l^i \leftarrow
\mathbf{z}_l^i/\|\mathbf{z}_l^i\|_2$
\STATE $\hat{\mathbf{z}}_v^i \leftarrow
\mathbf{z}_v^i/\|\mathbf{z}_v^i\|_2,\quad
\hat{\mathbf{e}}_l^i \leftarrow
\mathbf{e}_l^i/\|\mathbf{e}_l^i\|_2$
\STATE $\mathbf{S}_{1,ij} \leftarrow
\hat{\mathbf{z}}_v^{i\top}\hat{\mathbf{e}}_l^j$
\COMMENT{linguistic space}
\STATE $\mathbf{S}_{2,ij} \leftarrow
\hat{\mathbf{e}}_v^{i\top}\hat{\mathbf{z}}_l^j$
\COMMENT{visual space}
\STATE $\tau \leftarrow \exp(-\ell)$
\STATE $\mathcal{L}_1 \leftarrow
\mathcal{L}_{\mathrm{NCE}}(\mathbf{S}_1;\tau)$
\STATE $\mathcal{L}_2 \leftarrow
\mathcal{L}_{\mathrm{NCE}}(\mathbf{S}_2;\tau)$
\STATE $\mathcal{L} \leftarrow
\frac{1}{2}(\mathcal{L}_1+\mathcal{L}_2)$
\RETURN $\mathcal{L}$
\end{algorithmic}
\end{algorithm}

\subsection{Training and Evaluation Procedures}
\label{app:impl_projection}

Algorithm~\ref{alg:forward} and Algorithm~\ref{alg:eval} describe the
training and evaluation procedures, respectively. During training, the
forward pass computes contrastive losses in two complementary
subspaces: one pairs raw image features with projected text
representations in $\mathbb{R}^{d_v}$, while the other pairs projected
image features with raw text representations in $\mathbb{R}^{d_l}$.
At evaluation time, the model trained on synthetic data constructs a
fused similarity matrix by averaging scores from both projection
directions, and retrieval metrics are computed from this fused matrix
through top-$k$ selection.

\begin{algorithm}[h]
\caption{Similarity Matrix Construction with Asymmetric Dual Projection at Evaluation}
\label{alg:eval}
\begin{algorithmic}[1]
\REQUIRE Test images $\{\mathbf{x}_v^i\}_{i=1}^{N_I}$, test texts $\{\mathbf{x}_l^j\}_{j=1}^{N_T}$, trained model $(f_v, f_l, \mathbf{W}_l^{\uparrow}, \mathbf{W}_v^{\downarrow}, \tau)$
\ENSURE Score matrices $\mathbf{S}^{i2t} \in \mathbb{R}^{N_I \times N_T}$, $\mathbf{S}^{t2i} \in \mathbb{R}^{N_T \times N_I}$

\STATE $\hat{\mathbf{e}}_v^i,\, \hat{\mathbf{z}}_v^i \leftarrow \mathrm{Norm}(f_v(\mathbf{x}_v^i)),\; \mathrm{Norm}(\mathbf{W}_v^{\downarrow} f_v(\mathbf{x}_v^i)), \quad \forall\, i$
\STATE $\hat{\mathbf{e}}_l^j,\, \hat{\mathbf{z}}_l^j \leftarrow \mathrm{Norm}(f_l(\mathbf{x}_l^j)),\; \mathrm{Norm}(\mathbf{W}_l^{\uparrow} f_l(\mathbf{x}_l^j)), \quad \forall\, j$

\STATE $\mathbf{R}_{ij} \leftarrow \frac{1}{2\tau}\bigl(\hat{\mathbf{e}}_v^{i\top} \hat{\mathbf{z}}_l^j + \hat{\mathbf{z}}_v^{i\top} \hat{\mathbf{e}}_l^j\bigr)$

\STATE $\mathbf{S}^{i2t} \leftarrow \mathrm{TopKMask}(\mathbf{R}, k)$
\STATE $\mathbf{S}^{t2i} \leftarrow \mathrm{TopKMask}(\mathbf{R}^{\top}, k)$

\RETURN $\mathbf{S}^{i2t}$, $\mathbf{S}^{t2i}$
\end{algorithmic}
\end{algorithm}

\section{Comparison Methods}
\label{app:baselines}

\paragraph{Coreset Selection.}
Random (Rand) uniformly samples image-text pairs i.i.d.\ from the full
dataset. This unbiased baseline is computationally cheap but ignores
sample informativeness, which limits performance at low compression
ratios. Herding (Herd)~\cite{welling2009herding} is a greedy algorithm
that iteratively selects the sample whose inclusion best approximates
the full dataset's feature mean, minimizing first-order distributional
discrepancy. K-Center (K-Cent)~\cite{farahani2009facility} is a
coverage-based strategy that iteratively picks the point farthest from
the current subset, solving an approximate $k$-center problem to
maximize geometric coverage in the feature space. Forgetting
(Forget)~\cite{tonevaempirical} counts how often each sample
transitions from correctly to incorrectly matched during training;
low-forgetting (easy) samples are discarded first, retaining the most
informative hard examples.

\paragraph{Dataset Distillation.}
LoRS~\cite{xu2024low} is a trajectory-matching multimodal distillation
method that applies low-rank factorization to the cross-modal similarity matrix, reducing memory and
computation while preserving key alignment information.
CovMatch~\cite{leecovmatch} is a distribution-matching method that
projects both modalities into a shared space and aligns the
cross-modal covariance between real and synthetic data, relying on a
fixed model initialization for optimization stability.
RepBlend~\cite{zhangbeyond} is a trajectory-matching method that freezes pretrained encoders and symmetrically matches the trajectories of image and text projection heads, while blending synthetic representations to mitigate modality collapse and enhance intra-modal diversity. PTM-ST~\cite{guomultimodal} is a staged distillation
framework with two components. PTM partitions the distillation process
into multiple phases with dynamically shifted trajectory-matching
intervals, enabling progressive cross-modal knowledge absorption. ST
replaces the original high-variance expert trajectory with a smoothed
linear interpolation between endpoints, weighted by cumulative
distance. Combined with EMA updates, this suppresses gradient
oscillation and bridges inter-stage knowledge gaps, improving both
distillation stability and retrieval performance.

\begin{table}[t]
  \centering
  \caption{Hyperparameter configurations of \textsc{MDA-DD} on
    Flickr30K and MS-COCO under varying synthetic budget sizes.}
  \label{tab:hyperparameters_v2}
  \begin{tabular}{lcccccc}
    \toprule
    \multirow{2}{*}{Hyperparameter}
      & \multicolumn{3}{c}{Flickr30K}
      & \multicolumn{3}{c}{MS-COCO} \\
    \cmidrule(r){2-4} \cmidrule(l){5-7}
    & 100 & 200 & 500 & 100 & 200 & 500 \\
    \midrule
    Scaling factor $\rho$
      & 2 & 2 & 1 & 2 & 2 & 1 \\
    Feature matching weight $\lambda$
      & 0.1 & 0.1 & 0.5 & 0.1 & 0.1 & 0.5 \\
    Batch size (real)
      & 128 & 128 & 128 & 128 & 128 & 128 \\
    Batch size (synthetic)
      & -- & -- & 200 & -- & -- & 200 \\
    Learning rate (image)
      & 1 & 1 & 1 & 1 & 1 & 1 \\
    Learning rate (text)
      & 1 & 1 & 1 & 1 & 1 & 1 \\
    Iterations
      & 1000 & 1000 & 1000 & 1000 & 1000 & 1000 \\
    Number of models in queue $K$
      & 8 & 8 & 8 & 8 & 8 & 8 \\
    Maximum model lifetime $T_{\max}$
      & 200 & 200 & 200 & 200 & 200 & 200 \\
    \bottomrule
  \end{tabular}
\end{table}
\section{Additional Ablation Studies}
\label{app:ablation}

\subsection{Component-Wise Contribution}
\label{app:ablation_component}

\paragraph{Baseline reproduction.}
To ensure a fair comparison, we re-run CovMatch
under identical hardware and optimization settings as \textsc{MDA-DD}.
All experiments in this ablation use a single NVIDIA RTX
5090 (32\,GB). We set the following CovMatch hyperparameters to match
those of our method and its ablated variants: inner-loop iterations
($50$), outer-loop iterations ($200$), synthetic batch size,
feature-matching weight $\lambda$, and scaling factor $\rho$.
The remaining architecture-level settings (image encoder, text encoder,
image resolution, and synthetic budget $N$) are also kept consistent
with the main experiments (Table~\ref{tab:hyperparameters_v2}). This
controlled setup isolates the effect of each proposed component,
namely the model-distribution-aware objective (\S\ref{sec:mda}),
asymmetric dual projection (\S\ref{sec:projection}), and the SOCM
statistic (\S\ref{sec:socm}), from confounding factors such as
hardware differences or hyperparameter tuning.

\subsection{Cross-Architecture Generalization}
\label{app:cross-architecture}

\paragraph{Network architectures.}
Table~\ref{tab:networks} lists all models used in our experiments.
The image encoders are NFNet, NF-ResNet, NF-RegNet, and ViT; the text
encoder is BERT throughout.

\begin{table}[t]
\centering
\caption{Network architectures and parameter counts.}
\label{tab:networks}
\begin{tabular}{llr}
\toprule
Architecture & Model Variant & Parameter Count \\
\midrule
NFNet~\cite{brock2021high} & nfnet\_l0 & 26.4M \\
NF-ResNet~\cite{brockcharacterizing} & nf\_resnet50 & 25.6M \\
NF-RegNet~\cite{xu2022regnet} & nf\_regnet\_b1 & 21.0M \\
ViT~\cite{dosovitskiyimage} & vit\_base\_patch16\_224 & 85.8M \\
BERT~\cite{devlin2019bert} & bert-base-uncased & 110.0M \\
\bottomrule
\end{tabular}
\end{table}

\begin{table}[t]
\centering
\caption{Cross-architecture evaluation on Flickr30K with $N=100$
  synthetic pairs. All methods are distilled with NFNet+BERT and
  evaluated on unseen image encoders.}
\label{tab:cross_arch_full}
\small
\setlength{\tabcolsep}{3.5pt}
\begin{tabular}{llcccccc}
\toprule
Backbone & Method & IR@1 & IR@5 & IR@10 & TR@1 & TR@5 & TR@10 \\
\midrule

\multirow{6}{*}{ResNet+BERT}
  & Random   & 2.2 & 8.0  & 13.4 & 3.0 & 10.1 & 16.4 \\
  & MTT-VL   & 1.8 & 7.0  & 11.9 & 2.9 & 10.5 & 16.4 \\
  & LoRS     & 1.6 & 7.0  & 12.1 & 2.9 & 11.1 & 18.0 \\
  & RepBlend & 1.2 & 5.3  & 9.7  & 4.1 & 11.2  & 18.9 \\
  & CovMatch & 4.7 & 15.4 & 23.4 & 5.5 & 17.3 & 26.6 \\
  & \textbf{Ours}
        & \textbf{13.3} & \textbf{35.4} & \textbf{47.3}
        & \textbf{13.6} & \textbf{33.8} & \textbf{47.0} \\
\cmidrule(lr){1-8}

\multirow{6}{*}{RegNet+BERT}
  & Random   & 2.2 & 7.6  & 12.4 & 2.6 & 9.7  & 15.3 \\
  & MTT-VL   & 1.5 & 6.0  & 10.6 & 2.9 & 9.2  & 15.0 \\
  & LoRS     & 1.6 & 6.4  & 11.2 & 2.5 & 11.0 & 17.7 \\
  & RepBlend & 1.8 & 6.8  & 11.1 & 3.2 & 14.0 & 21.0 \\
  & CovMatch & 3.9 & 13.8 & 21.7 & 5.6 & 16.8 & 26.0 \\
  & \textbf{Ours}
        & \textbf{10.8} & \textbf{29.6} & \textbf{42.7}
        & \textbf{12.2} & \textbf{29.9} & \textbf{43.1} \\
\cmidrule(lr){1-8}

\multirow{6}{*}{ViT+BERT}
  & Random   & 2.8 & 9.3  & 15.1 & 4.2 & 13.3 & 19.8 \\
  & MTT-VL   & 2.1 & 7.8  & 13.1 & 3.9 & 12.2 & 18.6 \\
  & LoRS     & 2.0 & 7.3  & 12.7 & 3.2 & 11.7 & 19.0 \\
  & RepBlend & 2.4 & 9.5  & 15.3 & 4.8 & 16.6 & 23.9 \\
  & CovMatch & 3.7 & 13.5 & 21.7 & 5.2 & 18.6 & 28.0 \\
  & \textbf{Ours}
        & \textbf{9.9}  & \textbf{30.6} & \textbf{43.8}
        & \textbf{13.0} & \textbf{33.7} & \textbf{45.9} \\

\bottomrule
\end{tabular}
\end{table}

\paragraph{Cross-architecture evaluation.}
Table~\ref{tab:cross_arch_full} reports the full cross-architecture
results on Flickr30K with 100 synthetic pairs. All methods perform
distillation using NFNet+BERT and are evaluated on unseen image
encoders (NF-ResNet50, NF-RegNet-B1, ViT-B/16) while keeping the text
encoder fixed. Trajectory-based methods (MTT-VL, LoRS, RepBlend)
generally perform below random selection on image-retrieval metrics,
indicating limited transferability across visual encoders. In contrast,
\textsc{MDA-DD} consistently outperforms the compared methods across all
six retrieval metrics, achieving approximately $2.7$--$2.8\times$ the
IR@1 of CovMatch across the three unseen architectures. These results
demonstrate improved cross-architecture retrieval performance with
synthetic data distilled using \textsc{MDA-DD}.

\subsection{Sensitivity Analysis}
\label{app:ablation_study}

\begin{figure}[t]
  \centering
  \includegraphics[width=1\textwidth, trim=0 0 0 0, clip]{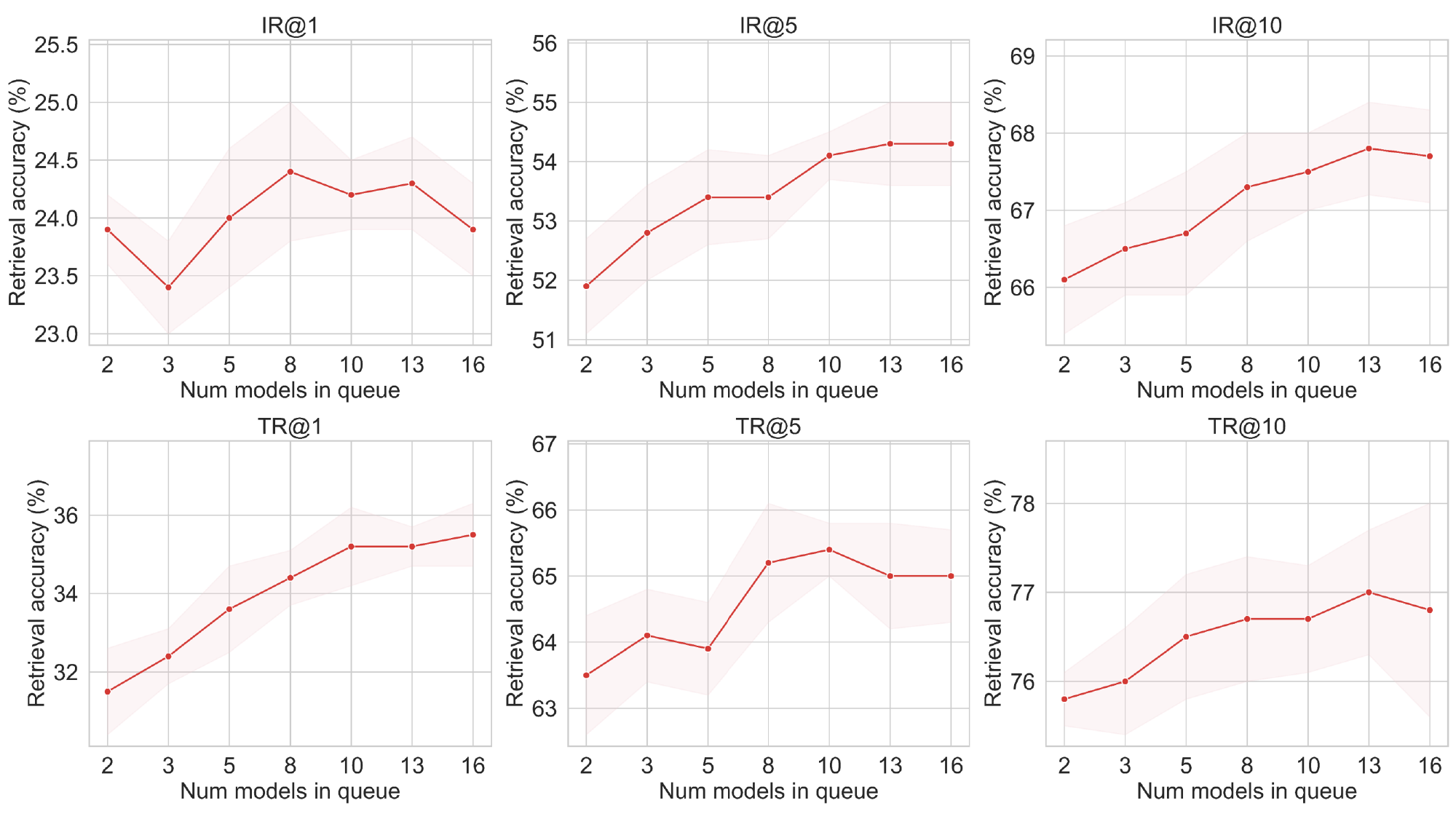}
  \caption{Effect of queue size $K$ on cross-modal retrieval
    performance (Flickr30K, 100 synthetic pairs). Performance improves
    steadily up to $K\!=\!8$ and saturates beyond $K\!=\!10$.}
  \label{fig:queue_size}
  \vspace{-5pt}
\end{figure}

\paragraph{Effect of queue size $K$.}
Figure~\ref{fig:queue_size} reports retrieval performance on Flickr30K
with 100 synthetic pairs as a function of the queue size
($K \in \{2, 3, 5, 8, 10, 13, 16\}$), with all other hyperparameters
held fixed. All six metrics improve consistently as $K$ increases from
2 to 8, with further marginal gains at $K\!=\!10$ on certain metrics
such as IR@5 and TR@1. This trend agrees with the theoretical
expectation from Appendix~\ref{app:proof_particle}: a larger queue
improves coverage of the model-state distribution $P(\Theta)$, thereby
narrowing the approximation gap between the particle objective
(Eq.~\ref{eq:particle_obj}) and the distributional objective
(Eq.~\ref{eq:dist_obj}). Beyond $K\!=\!10$, performance saturates or
slightly decreases on several metrics, indicating diminishing returns
from additional particles. Since $K\!=\!8$ and $K\!=\!10$ yield
comparable accuracy while larger queues incur proportionally higher
GPU memory and per-iteration cost, we adopt $K\!=\!8$ as the default
to balance performance and efficiency.

\begin{figure}[t]
  \centering
  \includegraphics[width=1\textwidth, trim=0 0 0 0, clip]{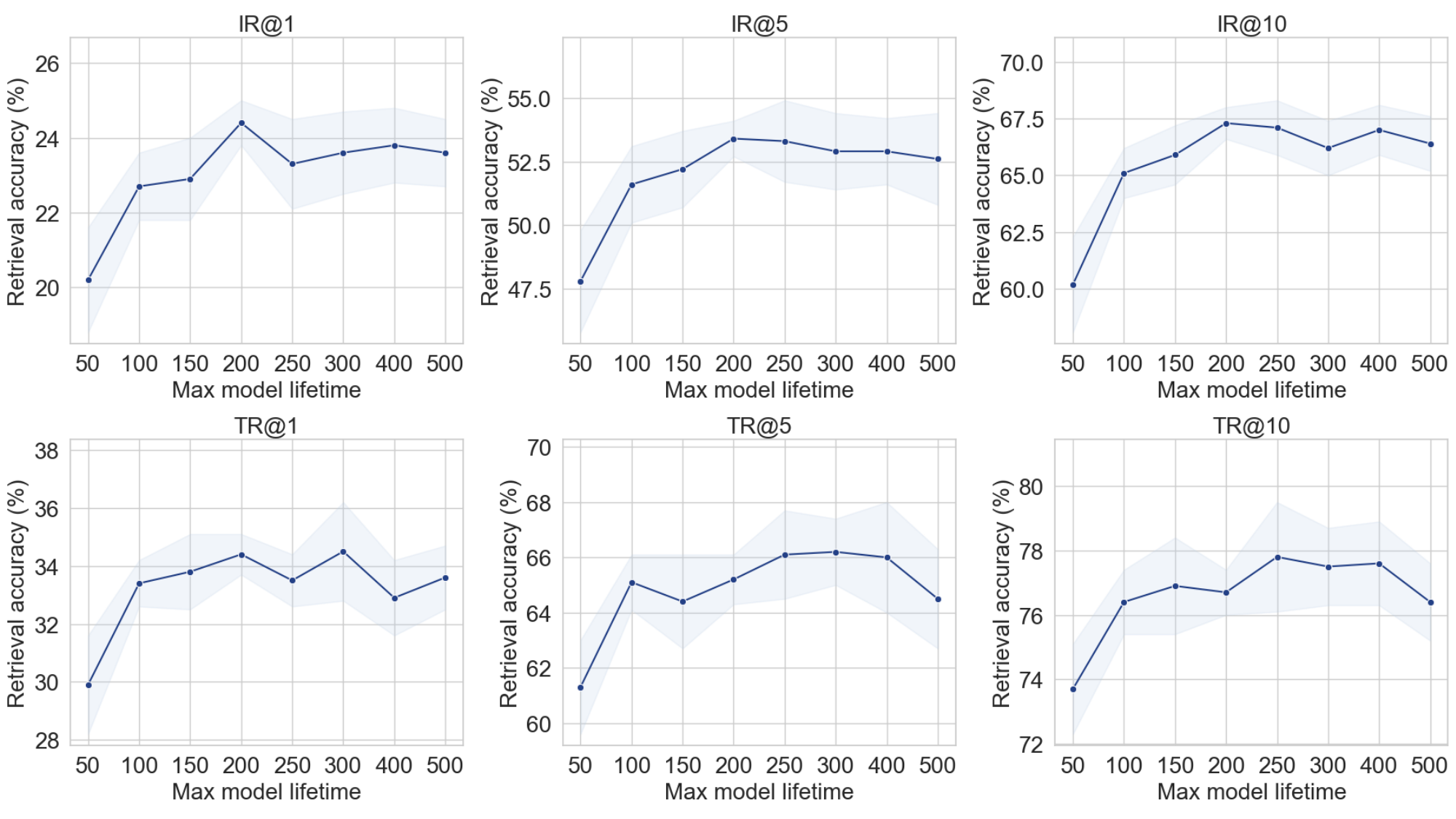}
  \caption{Effect of model lifetime $T_{\max}$ on cross-modal
    retrieval performance (Flickr30K, 100 synthetic pairs). The
    optimum lies around $T_{\max}\!=\!200$--$250$.}
  \label{fig:lifetime}
\end{figure}

\paragraph{Effect of model lifetime $T_{\max}$.}
Figure~\ref{fig:lifetime} shows retrieval performance on Flickr30K
with 100 synthetic pairs as $T_{\max}$ varies from 50 to 500.
Performance rises sharply from 50 to 200, peaks around
$T_{\max}\!=\!200$--$250$ across most metrics (e.g., IR@1 reaches
24.5\% at 200), and plateaus or slightly declines beyond 300. Too
short a lifetime limits the training-stage coverage of each particle,
reducing the diversity of model states in the queue. Too long a
lifetime causes particles to drift into late-training regimes where
the feature geometry changes slowly, diminishing the informational
gain from queue refreshes. We set $T_{\max}\!=\!200$ as the default,
as it provides sufficient training-stage diversity while maintaining
frequent queue renewal.
\section{Distillation under Alternative Budget Settings}
\label{app:distill_other_pairs}

\begin{table}[t]
  \centering
  \caption{Cross-modal retrieval performance with 50 synthetic pairs
    (0.2\% of Flickr30K, 0.4‰ of MS-COCO). Results are reported as
    mean $\pm$ standard deviation over 5 independent runs.}
  \label{tab:distill_50}
  \small
  \setlength{\tabcolsep}{3.5pt}
  \begin{tabular}{llllcccccc}
    \toprule
    Dataset & Pairs & Ratio & Method
      & IR@1 & IR@5 & IR@10 & TR@1 & TR@5 & TR@10 \\
    \midrule
    \multirow{3}{*}{Flickr30K}
      & \multirow{3}{*}{50}
      & \multirow{3}{*}{0.2\%}
      & RepBlend
        & 7.5{\tiny$\pm$0.5} & 24.8{\tiny$\pm$1.1} & 36.0{\tiny$\pm$1.7}
        & 12.5{\tiny$\pm$0.5} & 34.7{\tiny$\pm$0.7} & 48.4{\tiny$\pm$0.8} \\
    & & & CovMatch
        & 7.8{\tiny$\pm$0.2} & 23.8{\tiny$\pm$0.5} & 35.0{\tiny$\pm$0.3}
        & 11.5{\tiny$\pm$0.3} & 31.8{\tiny$\pm$1.3} & 44.2{\tiny$\pm$1.1} \\
    & & & \textbf{Ours}
        & \textbf{20.5}{\tiny$\pm$1.0} & \textbf{49.4}{\tiny$\pm$0.8} & \textbf{63.8}{\tiny$\pm$0.6}
        & \textbf{30.6}{\tiny$\pm$0.5} & \textbf{60.0}{\tiny$\pm$0.7} & \textbf{71.8}{\tiny$\pm$1.0} \\
    \midrule
    \multirow{3}{*}{MS-COCO}
      & \multirow{3}{*}{50}
      & \multirow{3}{*}{0.4‰}
      & RepBlend
        & 3.0{\tiny$\pm$0.2} & 10.9{\tiny$\pm$0.4} & 18.6{\tiny$\pm$0.6}
        & 4.4{\tiny$\pm$0.2} & 15.6{\tiny$\pm$0.3} & 24.7{\tiny$\pm$0.4} \\
    & & & CovMatch
        & 2.3{\tiny$\pm$0.1} & 8.6{\tiny$\pm$0.2} & 14.8{\tiny$\pm$0.4}
        & 2.8{\tiny$\pm$0.2} & 10.3{\tiny$\pm$0.4} & 16.8{\tiny$\pm$0.6} \\
    & & & \textbf{Ours}
        & \textbf{5.0}{\tiny$\pm$0.4} & \textbf{16.5}{\tiny$\pm$0.8} & \textbf{26.0}{\tiny$\pm$1.1}
        & \textbf{7.7}{\tiny$\pm$0.3} & \textbf{22.7}{\tiny$\pm$0.9} & \textbf{33.8}{\tiny$\pm$1.0} \\
    \bottomrule
  \end{tabular}
\end{table}

\paragraph{Extreme compression with 50 synthetic pairs.}
Table~\ref{tab:distill_50} evaluates performance under an extreme
compression setting using only 50 synthetic image-text pairs (0.2\% of
Flickr30K, 0.4‰ of MS-COCO). At this budget, both RepBlend and
CovMatch produce comparable yet modest results on Flickr30K (7--8
IR@1). \textsc{MDA-DD}, by contrast, reaches 20.5 IR@1 and 30.6 TR@1,
outperforming the strongest baseline by roughly $2.6\times$ and
$2.4\times$, respectively. On MS-COCO, where the compression ratio is
even more extreme, \textsc{MDA-DD} achieves 1.67× and 2.17× the IR@1 of RepBlend and CovMatch, respectively. The
widening gap at lower budgets suggests that the
model-distribution-aware objective is particularly effective when each
synthetic sample must carry a larger share of the overall training
signal.

\newpage
\section{Visualization of Distilled Data}
\noindent
\begin{minipage}{\textwidth}
  \centering
  \includegraphics[width=0.98\textwidth, trim=0 0 0 0, clip]{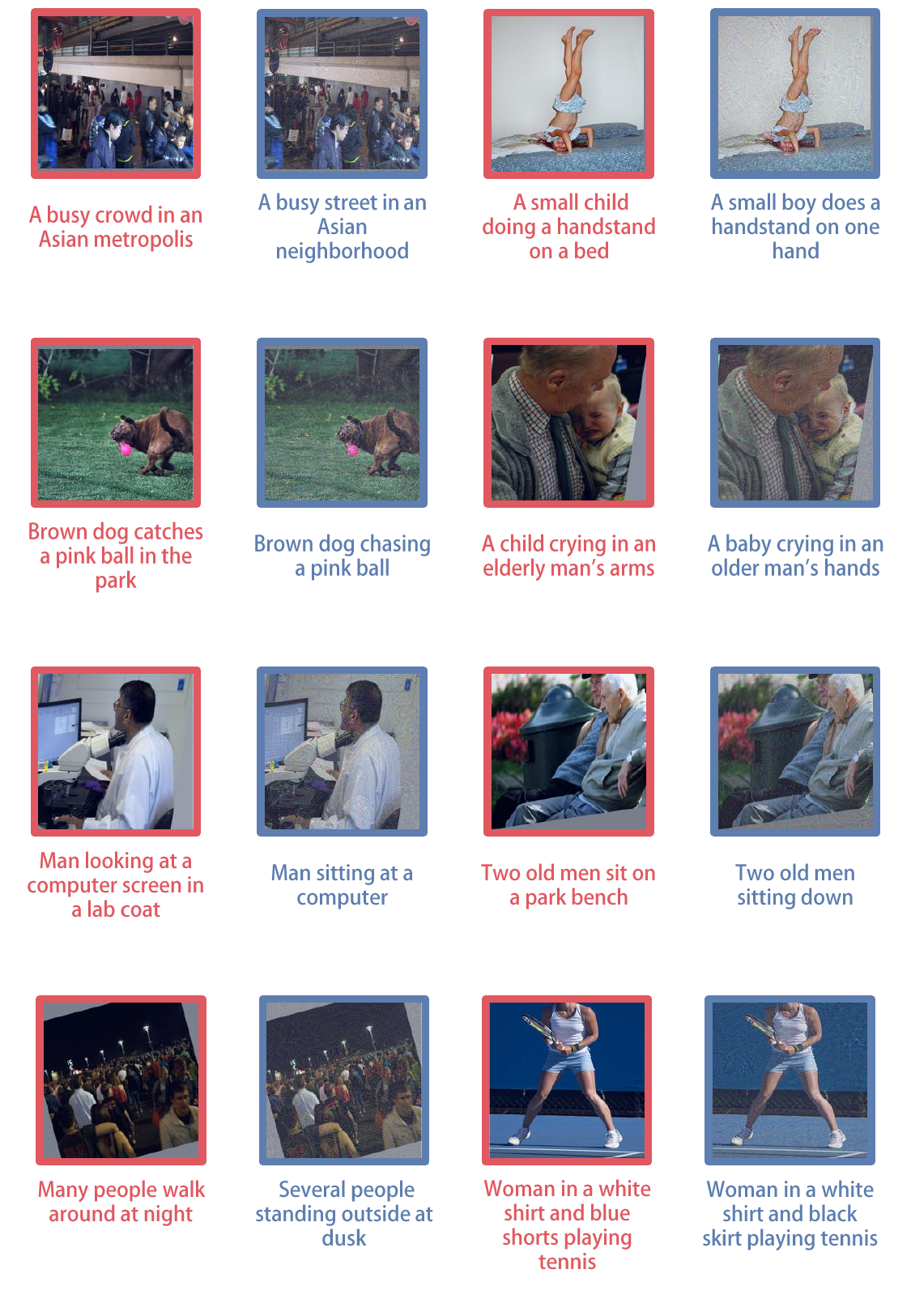}
  \captionof{figure}{Synthetic image-text pairs on Flickr30K before and after distillation. \textcolor[HTML]{E15B62}{Red (left):} original initialized pairs. \textcolor[HTML]{6484B3}{Blue (right):} distilled pairs produced by \textsc{MDA-DD}.}
  \label{fig:flickr_distilled}
\end{minipage}

\vspace{1em}

\noindent
\begin{minipage}{\textwidth}
  \centering
  \includegraphics[width=1\textwidth]{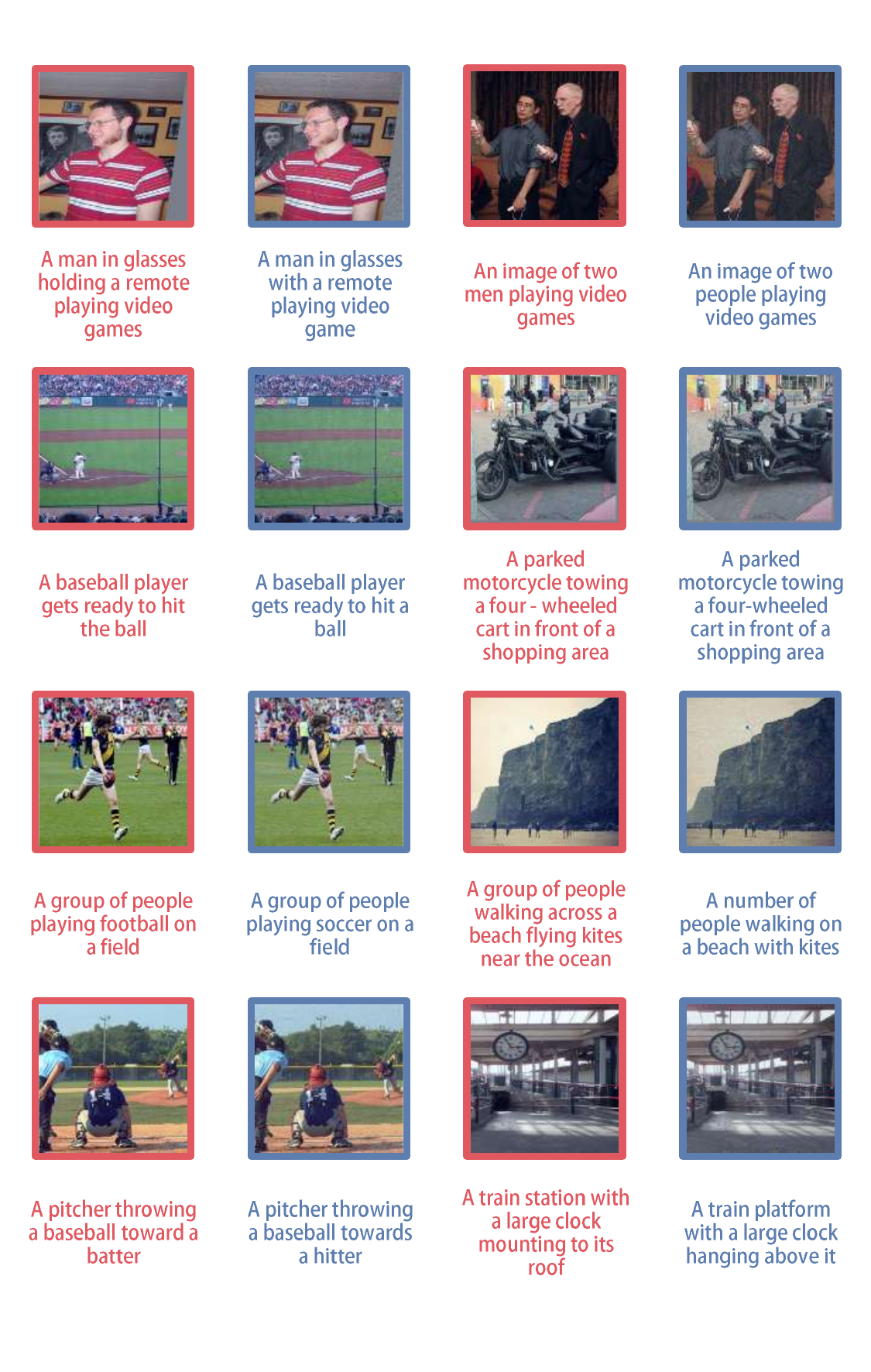}
  \captionof{figure}{Synthetic image-text pairs on MS-COCO before and after distillation. \textcolor[HTML]{E15B62}{Red (left)}: original initialized pairs. \textcolor[HTML]{6484B3}{Blue (right)}: distilled pairs produced by \textsc{MDA-DD}.}
  \label{fig:coco_distilled}
\end{minipage}

\end{document}